\documentclass[article,nojss]{jss}

\usepackage{graphicx}
\usepackage{amsmath}
\usepackage{amssymb}
\usepackage{enumerate}
\usepackage{subcaption}
\usepackage{algorithm}
\usepackage{algorithmicx}
\usepackage{algpseudocode}

\author{Adam Kapelner \\ Queens College, \\  City University of New York \And 
            Justin Bleich \\     The Wharton School of the \\ University of Pennsylvania}
\title{\pkg{bartMachine}: Machine Learning with Bayesian Additive Regression Trees}

\Plainauthor{Adam Kapelner, Justin Bleich} 
\Plaintitle{bartMachine: Machine Learning with Bayesian Additive Regression Trees} 
\Shorttitle{\pkg{bartMachine}: Machine Learning with Bayesian Additive Regression Trees} 

\Abstract{
We present a new package in \proglang{R} implementing Bayesian additive regression trees (BART). The package introduces many new features for data analysis using BART such as variable selection, interaction detection, model diagnostic plots, incorporation of missing data and the ability to save trees for future prediction. It is significantly faster than the current \proglang{R} implementation, parallelized, and capable of handling both large sample sizes and high-dimensional data. 
}
\Plainkeywords{Bayesian, machine learning, statistical learning, non-parametric, R, Java,} 

\Address{
  Adam Kapelner\\
  Department of Mathematics \\
  Queens College,  City University of New York \\
  64-19 Kissena Blvd Room 325 \\
  Flushing, NY, 11367 \\
  E-mail: \email{kapelner@qc.cuny.edu}\\
  URL: \url{http://scholar.google.com/citations?user=TzgMmnoAAAAJ}
}

\newcommand{\bv}[1]{\boldsymbol{#1}}
\newcommand{\qu}[1]{``#1''}
\renewcommand{\min}[1]{\text{min}\braces{#1}}
\renewcommand{\max}[1]{\text{max}\braces{#1}}
\newcommand{\treet}[1]{\mathcal{T}_{#1}}
\newcommand{\treeleaft}[1]{\treet{#1}^{\mathcal{M}}}
\newcommand{\leaf}{\mathcal{M}}
\newcommand{\beqn}{\begin{eqnarray*}}
\newcommand{\eeqn}{\end{eqnarray*}}
\newcommand{\bneqn}{\begin{eqnarray}}
\newcommand{\eneqn}{\end{eqnarray}}
\newcommand{\parens}[1]{\left(#1\right)}
\newcommand{\braces}[1]{\left\{#1\right\}}
\newcommand{\bracks}[1]{\left[#1\right]}
\newcommand{\squared}[1]{\parens{#1}^2}
\newcommand{\tothepow}[2]{\parens{#1}^{#2}}
\newcommand{\errorrv}{\mathcal{E}}
\newcommand{\berrorrv}{\bv{\errorrv}}
\newcommand{\normnot}[2]{\mathcal{N}\parens{#1,\,#2}}
\newcommand{\multnormnot}[3]{\mathcal{N}_{#1}\parens{#2,\,#3}}
\newcommand{\invgammanot}[2]{\text{InvGamma}\parens{#1,\,#2}}
\newcommand{\uniform}[2]{\mathrm{U}\parens{#1,\,#2}}
\newcommand{\zerovec}{\bv{0}}
\newcommand{\0}{\zerovec}
\newcommand{\sigsq}{\sigma^2}
\newcommand{\I}{\bv{I}}
\newcommand{\X}{\bv{X}}
\newcommand{\R}{\bv{R}}
\newcommand{\Z}{\bv{Z}}
\newcommand{\Y}{\bv{Y}}
\newcommand{\M}{\bv{M}}
\newcommand{\x}{\bv{x}}
\newcommand{\y}{\bv{y}}
\newcommand{\prob}[1]{\mathbb{P}\parens{#1}}
\newcommand{\cprob}[2]{\prob{#1~|~#2}}
\newcommand{\expe}[1]{\mathbb{E}\bracks{#1}}
\newcommand{\expesub}[2]{\mathbb{E}_{#1}\bracks{#2}}
\newcommand{\cexpe}[2]{\expe{#1 ~ | ~ #2}}
\newcommand{\iid}{~{\buildrel iid \over \sim}~}
\newcommand{\mathand}{~~\text{and}~~}
\newcommand{\yhat}{\hat{y}}
\newcommand{\oneover}[1]{\frac{1}{#1}}
\newcommand{\Hint}{H_{\text{internals}}}
\newcommand{\Hterminals}{H_{\text{terminals}}}

\newcommand{\reals}{\mathbb{R}}
\newcommand{\padjeta}{p_{\text{adj}}(\eta)}
\newcommand{\padjetastar}{p_{\text{adj}}(\eta^*)}
\newcommand{\nadjeta}{n_{j\cdot\text{adj}}(\eta)}
\newcommand{\nadjetastar}{n_{j^*\cdot\text{adj}}(\eta^*)}

\newcommand{\probsplit}[1]{\mathbb{P}_{\text{SPLIT}}\parens{#1}}
\newcommand{\probrule}[1]{\mathbb{P}_{\text{RULE}}\parens{#1}}
\newcommand{\Roneton}{R_1, \ldots, R_n}
\newcommand{\Rlonetonl}{R_{\ell_1}, \ldots, R_{\ell_{n_\ell}}}
\newcommand{\RLlonetonlL}{R_{\ell_{L,1}}, \ldots, R_{\ell_{L, n_{\ell,L}}}}
\newcommand{\RRlonetonlR}{R_{\ell_{R,1}}, \ldots, R_{R,\ell_{n_{\ell,R}}}}
\newcommand{\Rbar}{\bar{R}}
\newcommand{\sigsqmu}{\sigsq_\mu}
\newcommand{\doneover}[1]{\dfrac{1}{#1}}
\newcommand{\Rones}{R_{1,1}, \ldots, R_{1, \none}}
\newcommand{\Rtwos}{R_{2,1}, \ldots, R_{2, \ntwo}}
\newcommand{\Ronestars}{R_{1^*,1}, \ldots, R_{1^*, \nonestar}}
\newcommand{\Rtwostars}{R_{2^*,1}, \ldots, R_{2^*, \ntwostar}}
\newcommand{\none}{n_{1}}
\newcommand{\ntwo}{n_{2}}
\newcommand{\nonestar}{n_{1^*}}
\newcommand{\ntwostar}{n_{2^*}}
\newcommand{\etastar}{\eta_*}
\newcommand{\etaone}{\eta_1}
\newcommand{\etatwo}{\eta_2}
\newcommand{\etaonestar}{\eta_{1^*}}
\newcommand{\etatwostar}{\eta_{2^*}}
\renewcommand{\exp}[1]{\text{exp}\parens{#1}}

\begin{document}

\section{Introduction}

Ensemble-of-trees methods have become popular choices for forecasting in both regression and classification problems. Algorithms such as random forests \citep{Breiman2001} and stochastic gradient boosting \citep{Friedman2002} are two well-established and widely employed procedures. Recent advances in ensemble methods include dynamic trees \citep{Taddy2011} and Bayesian additive regression trees \citep[BART,][]{Chipman2010}, which depart from predecessors in that they rely on an underlying Bayesian probability model rather than a pure algorithm. BART has demonstrated substantial promise in a wide variety of simulations and real world applications such as predicting avalanches on mountain roads \citep{Blattenberger2014}, predicting how transcription factors interact with DNA \citep{Zhou2008} and predicting movie box office revenues \citep{Eliashberg2010}. This paper introduces \pkg{bartMachine}, a new \proglang{R} \citep{Rlang} package available from the Comprehensive \proglang{R} Archive Network at \url{http://CRAN.R-project.org/package=bartMachine} that significantly expands the capabilities of using BART for data analysis. 

Currently, there exists one other implementation of BART on CRAN: \pkg{BayesTree} \citep{BayesTree}, the package developed by the algorithm's original authors. One of the major drawbacks of this implementation is its lack of a \code{predict} function. Test data must be provided as an argument during the training phase of the model. Hence it is impossible to generate forecasts on future data without re-fitting withe entire model. Since the run time is not trivial, forecasting becomes an arduous exercise. A significantly faster implementation of BART that contains master-slave parallelization exists as \citet{Pratola2013}, but this is only available as standalone C++ source code and not integrated with \proglang{R}. Additionally, a recent package \proglang{dbarts} allows updating of BART with new predictors and response values to incorporate BART into a larger Bayesian model. \proglang{dbarts} relies on \proglang{BayesTree} as it's BART engine.

The goal of \pkg{bartMachine} is to provide a fast, easy-to-use, visualization-rich machine learning package for \proglang{R} users. Our implementation of BART is in \proglang{Java} and is integrated into \proglang{R} via \pkg{rJava} \citep{rJava}. From a runtime perspective, our algorithm is significantly faster and is parallelized, allowing computation on as many cores as desired. Not only is the model construction itself parallelized, but the additional features such as prediction, variable selection, and many others can be divided across cores as well.  

Additionally, we include a variety of expanded and new features. We implement the ability to save trees in memory and provide convenience functions for prediction on test data. We also include plotting functions for both credible and predictive intervals and plots for visually inspecting convergence of BART's Gibbs sampler. We expand variable importance exploration to include permutation tests and interaction detection. We implement recently developed features for BART including a principled approach to variable selection and the ability to incorporate in prior information for covariates \citep{Bleich2014}. We also implement the strategy found in \citet{Kapelner2013} to incorporate missing data during training and handle missingness during prediction.  Table~\ref{tab:bartcomp} emphasizes the differences in features between \proglang{bartMachine} and \proglang{BayesTree}, the two existing \proglang{R} implementations of BART.

\begin{table}[htp]
\centering
\begin{tabular}{rcc}
  \hline
Feature & \pkg{bartMachine} & \pkg{BayesTree} \\ 
  \hline
Implementation Language &  \proglang{Java} &  C++ \\
External Predict Function & Yes & No \\
Model Persistance Across Sessions & Yes & No \\ 
Parallelization & Yes & No \\ 
Native Missing Data Mechanism & Yes & No \\
Built-in Cross-Validation & Yes & No \\
Variable Importance & Statistical Tests & Exploratory \\
Tree Proposal Types & 3 Types & 4 Types \\
Partial Dependence Plots & Yes & Yes \\
Convergence Plots & Assess trees and $\sigsq$ & Assess $\sigsq$ \\
Model Diagnostics & Yes & No \\
Incorporation into Larger Model & No & Through \proglang{dbarts} \\ 
\hline
\end{tabular}
\caption{Comparison of features between \proglang{bartMachine} and \proglang{BayesTree}.}
\label{tab:bartcomp}
\end{table}

In Section~\ref{sec:background}, we provide an overview of BART with a special emphasis on the features that have been extended. In Section~\ref{sec:package} we provide a general introduction to the package, highlighting the novel features. Section~\ref{sec:regression_features} provides step-by-step examples of the regression capabilities and Section~\ref{sec:classification_features} introduces additional step-by-step examples of features unique to classification problems. We conclude in Section \ref{sec:discussion}. Appendix~\ref{app:implementation} discusses the details of our algorithm implementation and how it differs from \pkg{BayesTree}. Appendix~\ref{app:bakeoff} offers predictive performance comparisons. \pagebreak

\section{Overview of BART}\label{sec:background}

BART is a Bayesian approach to nonparametric function estimation using regression trees. Regression trees rely on recursive binary partitioning of predictor space into a set of hyperrectangles in order to approximate some unknown function $f$. Predictor space has dimension of the number of variables, which we denote $p$. Tree-based regression models have an ability to flexibly fit interactions and nonlinearities. Models composed of sums of regression trees have an even greater ability than single trees to capture interactions and non-linearities as well as additive effects in $f$. 

BART can be considered a sum-of-trees ensemble, with a novel estimation approach relying on a fully Bayesian probability model.  Specifically, the BART model can be expressed as:

\bneqn\label{eq:bart}
\Y = f(\X) + \berrorrv \approx \treeleaft{1}(\X) + \treeleaft{2}(\X) + \ldots + \treeleaft{m}(\X) + \berrorrv, \quad\quad \berrorrv \sim \multnormnot{n}{\zerovec}{\sigsq\I_n}
\eneqn

where $\Y$ is the $n \times 1$ vector of responses, $\X$ is the $n \times p$ design matrix (the predictors column-joined), $\berrorrv$ is the $n \times 1$ vector of noise. Here we have $m$ distinct regression trees, each composed of a tree structure, denoted by $\treet{}$, and the parameters at the terminal nodes (also called leaves), denoted by $\leaf$. The two together, denoted as $\treeleaft{}$ represents an entire tree with both its structure and set of leaf parameters.

The structure of a given tree $\treet{t}$ includes information on how any observation recurses down the tree. For each nonterminal (internal) node of the tree, there is a \qu{splitting rule} taking the form $\x_j < c$ consisting of the \qu{splitting variable} $\x_j$ and the \qu{splitting value} $c$. An observation moves to the left child node if the condition set by the splitting rule is satisfied and to the right child node otherwise. The process continues until a terminal node is reached. Then, the observation receives the leaf value of the terminal node. The sum of the $m$ leaf values becomes its predicted value. We denote the set of tree's leaf parameters as  $\leaf_t = \{\mu_{t,1}, \mu_{t,2}, \ldots, \mu_{t_{b_t}}\}$ where $b_t$ is the number of terminal nodes for a given tree.

BART can be distinguished from other ensemble-of-trees models due to its underlying probability model. As a Bayesian model, BART consists of a set of priors for the structure and the leaf parameters and a likelihood for data in the terminal nodes. The aim of the priors is to provide regularization, preventing any single regression tree from dominating the total fit. 

We provide an overview of the BART priors and likelihood and then discuss how draws from the posterior distribution are made. A more complete exposition can be found in \citet{Chipman2010}. 

\subsection{Priors and likelihood}\label{subsec:prior_likelihood}

The prior for the BART model has three components (1) the tree structure itself (2) the leaf parameters given the tree structure and (3) the error variance $\sigsq$ which is independent of the tree structure and leaf parameters:

\beqn 
\prob{\treeleaft{1},\ldots,\treeleaft{m}, \sigsq} &=& \bracks{\prod_{t}\prob{\treeleaft{t}}}\prob{\sigsq} 
   = \bracks{\prod_{t}\cprob{\leaf_t}{\treet{t}}\prob{\treet{t}}}\prob{\sigsq}  \\
&=& \bracks{\prod_{t}\prod_{\ell}\cprob{ \mu_{t,\ell}}{\treet{t}}\prob{\treet{t}}}\prob{\sigsq}
\eeqn

\noindent where the last line follows from an additional assumption of conditional independence of the leaf parameters given the tree's structure. 

We first describe $\prob{\treet{t}}$, the component of the prior which affects the locations of nodes within the tree. Nodes at depth $d$ are nonterminal with probability $\alpha(1+d)^{-\beta}$ where $\alpha \in (0,1)$ and $\beta \in [0, \infty]$. Depth is defined as distance from the root. Thus, the root itself has depth 0, its first child node has depth 1, etc. This prior form has the ability to enforce shallow tree structures, thereby limiting complexity of any single tree. Default values for these hyperparameters of $\alpha = 0.95$ and $\beta = 2$ are recommended by \citet{Chipman2010}. 

For nonterminal nodes, splitting rules occur in two parts. First, the predictor is randomly selected to serve as the splitting variable. In the original formulation, each available predictor is equally likely to be chosen from a discrete uniform distribution with probability that each varible selected is $1/p$. This is relaxed in our implementation to allow for a generalized Bernoulli distribution where the user specifies $p_1, p_2, \ldots, p_p$ (such that $\sum_{j=1}^p p_j = 1$), where each is a probability of variable selection. See \qu{covariate priors} (Section~\ref{subsec:cov_prior}) for further details. Note that \qu{structural zeroes,} variables that do not have any valid split values, are assigned probability zero (see Appendix \ref{subapp:grow_step} for details). Once the splitting variable is chosen, the splitting value is chosen from the multiset (the non-unique set) of available values at the node via the discrete uniform distribution. 

We now describe the prior component $\cprob{\leaf_t}{\treet{t}}$ which controls the leaf parameters. Given a tree with a set of terminal nodes, each terminal node (or leaf) has a continuous parameter (the leaf parameter) representing the \qu{best guess} of the response in this partition of predictor space. This parameter is the fitted value assigned to any observation that lands in that node. The prior on each of the leaf parameters is given as: $\mu_\ell \iid \normnot{\mu_\mu / m}{\sigsq_\mu}$. The expectation, $\mu_\mu$, is picked to be the range center, $(y_{\text{min}} + y_{\text{max}}) / 2$. The range center can be affected by outliers. If this is a concern, the user can log-transform the response or windsorize the response. 

The variance is empirically chosen so that the range center plus or minus $k = 2$ variances cover 95\% of the provided response values in the training set (where $k=2$ corresponding to 95\% coverage is only by default and can be customized). Thus, since there are $m$ trees, we are then automatically employing $\sigma_\mu$ such that $m \mu_\mu - k\sqrt{m} \sigma_\mu = y_{\text{min}}$ and  $m \mu_\mu + k\sqrt{m} \sigma_\mu = y_{\text{max}}$. The aim of this prior is to provide model regularization by shrinking the leaf parameters towards the center of the distribution of the response.

The final prior is on the error variance and is chosen to be $\sigsq \sim \invgammanot{\nu / 2}{\nu\lambda / 2}$. $\lambda$ is determined from the data so that there is a $q = 90\%$ a priori chance (by default) that the BART model will improve upon the RMSE from an ordinary least squares regression. Therefore, the majority of the prior probability mass lies below the RMSE from least squares regression. Additionally, this prior limits the probability mass placed on small values of $\sigsq$ to prevent overfitting.

Note that the adjustable hyperparameters are $\alpha$, $\beta$, $k$, $\nu$ and $q$. Default values generally provide good performance, but optimal tuning can be achieved via cross-validation, an automatic feature implemented and described in Section~\ref{subsec:model_building}.

Along with a set of priors, BART specifies the likelihood of responses in the terminal nodes. They are assumed a priori Normal with the mean being the \qu{best guess} in the leaf at the current moment (in the Gibbs sampler) and variance being the best guess of the variance at the moment i.e., $\y_\ell \sim \normnot{\mu_\ell}{\sigsq}$.

\subsection{Posterior distribution and prediction} \label{subsec:posterior}

A Gibbs sampler \citep{Geman1984} is employed to generate draws from the posterior distribution of $\mathbb{P}(\treeleaft{1}, \ldots, \treeleaft{m}, \sigsq ~|~\y)$. A key feature of the Gibbs sampler for BART is to employ a form of ``Bayesian backfitting'' \citep{Hastie2000} where the $j$th tree is fit iteratively, holding all other $m-1$ trees constant by exposing only the residual response that remains unfitted: 

\bneqn\label{eq:response_unfitted}
\R_{-j} := \y - \sum_{t \neq j} \treeleaft{t}(\X).
\eneqn

The Gibbs sampler,

\bneqn\label{eq:gibbs_sampler}
1: && \treet{1} ~|~ \R_{-1}, \sigsq \\ \nonumber
2: && \leaf_1 ~|~ \treet{1}, \R_{-1}, \sigsq \\ \nonumber
3: && \treet{2} ~|~ \R_{-2}, \sigsq \\ \nonumber
4: && \leaf_2 ~|~ \treet{2}, \R_{-2}, \sigsq \\  \nonumber
\vdots &&  \\\nonumber
2m -1: && \treet{m} ~|~ \R_{-m}, \sigsq \\\nonumber
2m: && \leaf_m ~|~ \treet{m}, \R_{-m}, \sigsq \\ \nonumber
2m + 1: && ~\,\sigsq ~|~ \treet{1}, \leaf_1, \ldots, \treet{m}, \leaf_m, \berrorrv, \nonumber
\eneqn

\noindent proceeds by first proposing a change to the first tree's structure $\treet{}$ which are accepted or rejected via a Metropolis-Hastings step \citep{Hastings1970}. Note that sampling from the posterior of the tree structure does not depend on the leaf parameters, as they can be analytically integrated out of the computation (see Appendix ~\ref{subapp:grow_step}). Given the tree structure, samples from the posterior of the $b$ leaf parameters $\leaf_1 := \braces{\mu_1, \ldots, \mu_b}$  are then drawn. This procedure proceeeds iteratively for each tree, using the updated set of partial residuals $\R_{-j}$. Finally, conditional on the updated set of tree structures and leaf parameters, a draw from the posterior of $\sigsq$ is made based on the full model residuals $\berrorrv := \y - \sum_{t = 1}^m \treeleaft{t}(\X)$.

Within a given terminal node, since both the prior and likelihood are normally distributed, the posterior of each of the leaf parameters in $\leaf$ is conjugate normal with its mean being a weighted combination of the likelihood and prior parameters (lines $2,~4, \ldots, 2m$ in Equation set \ref{eq:gibbs_sampler}). Due to the normal-inverse-gamma conjugacy, the posterior of $\sigsq$ is inverse gamma as well (line $2m+1$ in Equation set \ref{eq:gibbs_sampler}). The complete expressions for these posteriors can be found in \citet{Gelman2004}. 

Lines $1,~3,\ldots,~2m-1$ in Equation set \ref{eq:gibbs_sampler} rely on Metropolis-Hastings draws from the posterior of the tree distributions. These involve introducing small perturbations to the tree structure: growing a terminal node by adding two child nodes, pruning two child nodes (rendering their parent node terminal), or changing a split rule. We denote the three possible tree alterations as: GROW, PRUNE, and CHANGE.\footnote{In the original formulation, \citet{Chipman2010} include an additional alteration called SWAP. Due to the complexity of bookkeeping associated with this alteration, we do not implement it.} The mathematics associated with the Metropolis-Hastings step are tedious. Appendix~\ref{app:implementation} contains the explicit calculations. Once again, over many MCMC iterations, trees evolve to capture the fit left currently unexplained.

\citet{Pratola2013} argue that a CHANGE step is unnecessary for sufficient mixing of the Gibbs sampler. While we too observed this to be true for estimates of the posterior means, we found that omitting CHANGE can negatively impact the variable inclusion proportions (the feature introduced in Section~\ref{subsec:variable_importance}). As a result, we implement a modified CHANGE where we only propose new splits for nodes that are singly internal: both children nodes  are terminal nodes (details are given in Appendix~\ref{subapp:change_step}). After a singly internal node is selected we (1) select a new split attribute from the set of available predictors and (2) select a new split value from the multiset of avilable values (these two uniform splitting rules were explained in detail previously). We would like to emphasize that the CHANGE step does not alter tree structure.

All $2m+1$ steps represent a \textit{single} Gibbs iteration. We have observed that generally no more than 1,000 iterations are needed as ``burn-in'' (see Section~\ref{subsec:assumption_checking} for convergence diagnostics). An additional 1,000 iterations are usually sufficient to serve as draws from the posterior for $f(\x)$. A single predicted value $\hat{f}(\x)$ can be obtained by taking the average of the posterior values and a quantile estimate can be obtained by computing the appropriate quantile of the posterior values. Additional features of the posterior distribution will be discussed in Section~\ref{sec:regression_features}.

\subsection{BART for classification} \label{subsec:probit_bart}

BART can easily be modified to handle classification problems for categorical response variables. In \citet{Chipman2010},  only binary outcomes were explored but recent work has extended BART to the multiclass problem \citep{Kindo2013}. Our implementation handles binary classification and we plan to implement multiclass outcomes in a future release.

For the binary classification problem (coded with outcomes ``0'' and ``1''), we assume a probit model,

\beqn
\cprob{\Y = 1}{\X} = \Phi\parens{\treeleaft{1}(\X) + \treeleaft{2}(\X) + \ldots + \treeleaft{m}(\X)},
\eeqn

\noindent where $\Phi$ denotes the cumulative density function of the standard normal distribution. In this formulation, the sum-of-trees model serves as an estimate of the conditional probit at $\x$ which can be easily transformed into a conditional probability estimate of $Y=1$. 

In the classification setting, the prior on $\sigsq$ is not needed as the model assumes $\sigsq=1$. The prior on the tree structure remains the same as in the regression setting and a few minor modifications are required for the prior on the leaf parameters. 

Sampling from the posterior distribution is again obtained via Gibbs sampling with a Metropolis-Hastings step outlined in Section~\ref{subsec:posterior}. Following the data augmentation approach of \citet{Albert1993}, an additional vector of latent variables $\Z$ is introduced into the Gibbs sampler. Then, a new step is created in the Gibbs sampler where draws of $\Z\,|\,\y$ are obtained by conditioning on the sum-of-trees model:

\beqn
Z_i~|~y_i=1 &\sim& \max{N\parens{\sum_t \treeleaft{t}\parens{\X},1}, \,0} \mathand \\
Z_i~|~y_i = 0  &\sim& \min{N\parens{\sum_t \treeleaft{t}\parens{\X},1}, \,0}.
\eeqn

\noindent Next, $\Z$ is used as the response vector instead of $\y$ in all steps of  Equation~\ref{eq:gibbs_sampler}.

Upon obtaining a sufficient number of samples from the posterior, inferences can be made using the the posterior distribution of conditional probabilities and classification can be undertaken by applying a threshold to the to the means (or another summary) of these posterior probabilities. The relevant classification features of \pkg{bartMachine} are discussed in Section~\ref{sec:classification_features}.

\section[The bartMachine package]{The \pkg{bartMachine} package}\label{sec:package}

The package \pkg{bartMachine} provides a novel implementation of Bayesian additive regression trees in \proglang{R}. The algorithm is substantially faster than the current \proglang{R} package \pkg{BayesTree} and our implementation is parallelized at the MCMC iteration level during prediction.  Additionally, the interface with \pkg{rJava} allows for the entire posterior distribution of tree ensembles to persist throughout the \proglang{R} session, allowing for prediction and other calls to the trees without having to re-run the Gibbs sampler (a limitation in the current implementation). The model object cannot persist across sessions (using \proglang{R}'s save command for instance) and we view the addition of this feature as future work. Since our implementation is different from \pkg{BayesTree}, we provide a predictive accuracy bakeoff on different datasets in Appendix~\ref{app:bakeoff} which illustrates that the two are about equal.

\subsection{Speed improvements and parallelization}\label{subsec:speed_and_parallelization}

We make a number of significant speed improvements over the original implementation. 

First, \pkg{bartMachine} is fully parallelized (with the number of cores customizable) during model creation, prediction, and many of the other features. During model creation, we chose to parallelize by creating one independent Gibbs chain per core. Thus, if we employ the deafult 250 burn-in samples and 1,000 post burn-in samples and four cores, each core would sample 500 samples: 250 for a burn-in and 250 post burn-in samples. The final model will aggregate the post burn-in samples for the four cores yielding the desired 1,000 total burn-in samples. This has the drawback of effectively running the burn-in serially (which suffers from Amdahl's Law), but has the added benefit of reducing auto-correlation of the sum-of-trees samples in the posterior samples since the chains are independent which may provide greater predictive performance. Parallelization at the level of likelihood calculations is left for a future release as we were unable to address the costs of thread overhead. Parallelization for prediction and other features scale linearly in the number of cores.

Additionally, we take advantage of a number of additional computational shortcuts:

\begin{enumerate}[1.]
\item Computing the unfitted responses for each tree (Equation~\ref{eq:response_unfitted}) can be accomplished by keeping a running vector and making entry-wise updates as the Gibbs sampler (Equation~\ref{eq:gibbs_sampler}) progresses from step 1 to $2m$. Additionally, during the $\sigsq$ sampling (step $2m + 1$), the residuals do not have to be computed by dropping the data down all the trees.
\item Each node caches its acceptable variables for split rules and the acceptable unique split values so they do not need to be calculated at each tree sampling step. Recall from the discussion concerning uniform splitting rules in Section~\ref{subsec:prior_likelihood} that acceptable predictors and values change based on the data available at an arbitrary location in the tree structure. This speed enhancement, which we call \textit{memcache} comes at the expense of memory and may cause issues for large data sets. We include a toggle in our implementation defaulted to ``on.''
\item Careful calculations in Appendix~\ref{app:implementation} eliminate many unnecessary computations. For instance, the likelihood ratios are only functions of the squared sum of responses and no longer require computing the sum of the responses squared.
\end{enumerate}

\begin{figure}[htp]
\centering
\begin{subfigure}[c]{.48\textwidth}
                \centering
                \includegraphics[width=3.2in]{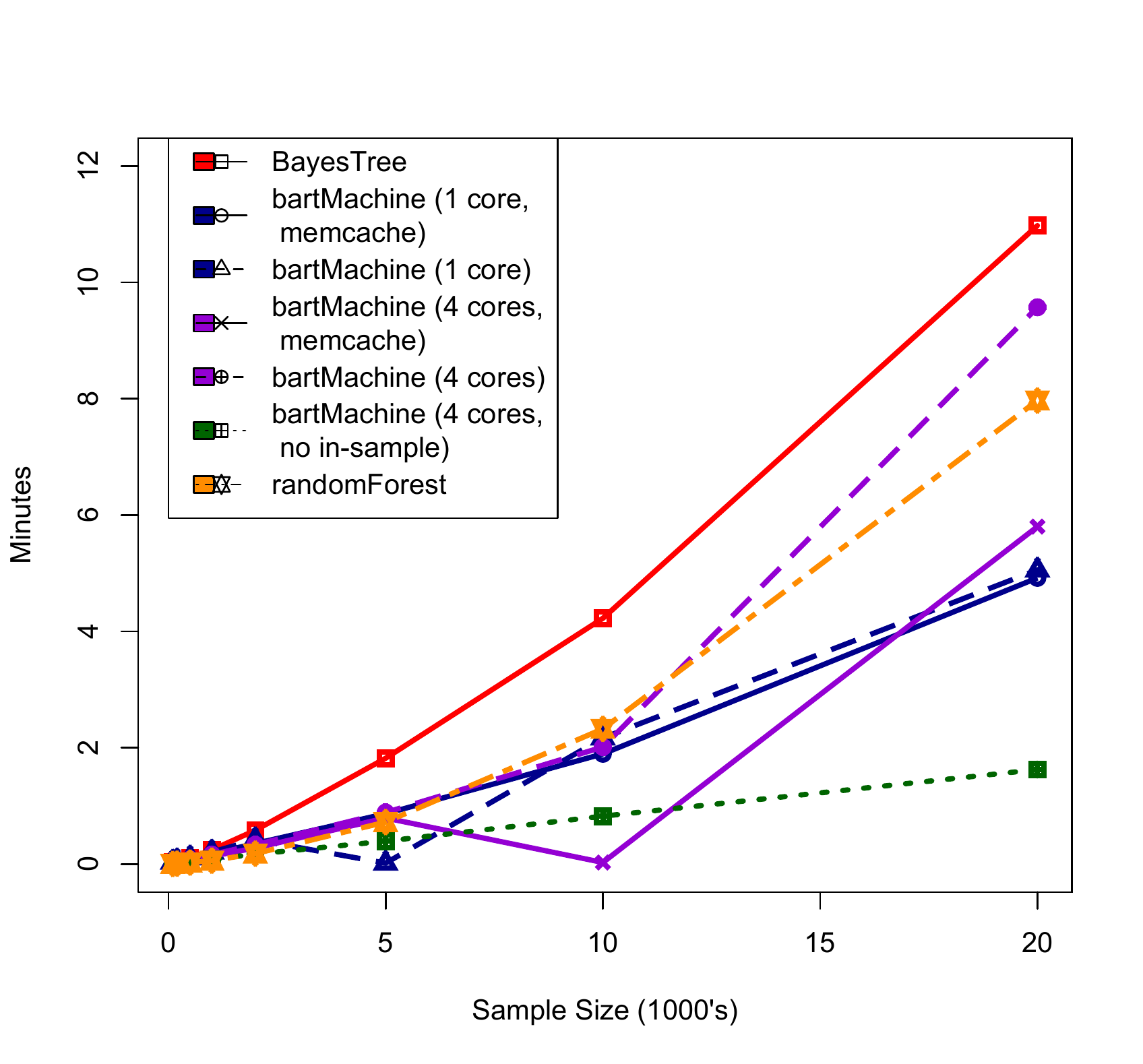}
                \caption{Large sample sizes}
                \label{fig:time_plots_large_sample}
        \end{subfigure}
\begin{subfigure}[c]{.48\textwidth}
                \centering
                \includegraphics[width=3.2in]{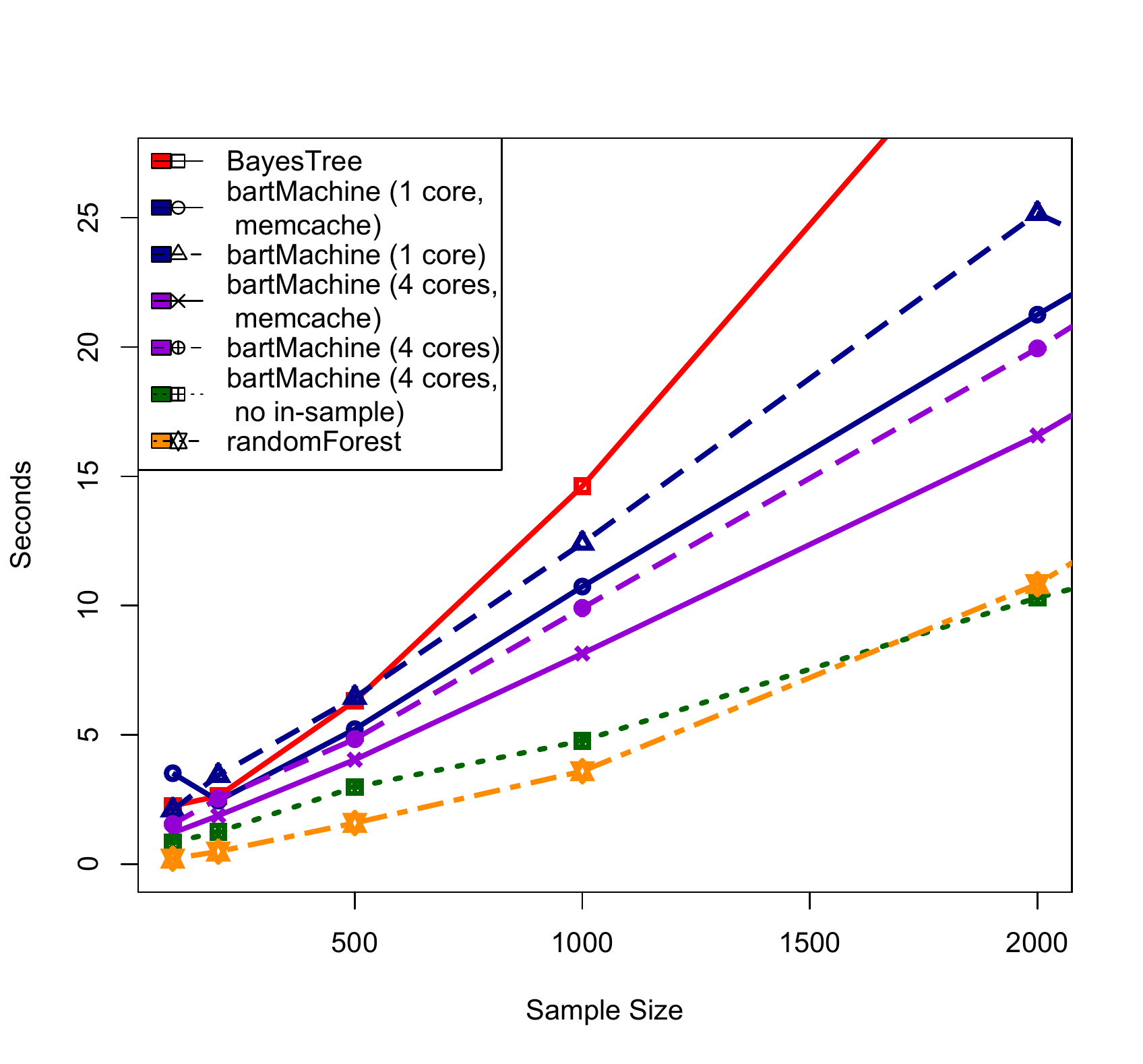}
                \caption{Small sample sizes}
                \label{fig:time_plots_small_sample}
        \end{subfigure}
\caption{Model creation times as a function of sample size for a number of settings of \pkg{bartMachine}, \pkg{BayesTree} and \texttt{RandomForests}. Simulations were run on a quad-core 3.4GHz Intel i5 desktop with 24GB of RAM running the Windows 7 64bit operating system.}
\label{fig:time_plots}
\end{figure}

Figure~\ref{fig:time_plots} displays model creation speeds for different values of $n$ on a linear regression model with $p=20$, normally distributed covariates, $\beta_1, \ldots, \beta_{20} \iid \uniform{-1}{1}$, and standard normal noise. Note that we do not vary $p$ as it was already shown in \citet{Chipman2010} that BART's computation time is largely unaffected by the dimensionality of the problem (relative to the influence of sample size).  We include results for BART using \pkg{BayesTree}, \pkg{bartMachine} with one and four cores, the \textit{memcache} option on and off, as well as four cores, \textit{memcache} off and computation of in-sample statistics off (all with $m=50$ trees). In-sample statistics by default are computing in-sample predictions ($\yhat$), residuals ($e := y - \yhat$), $L1$ error which is defined as $\sum_{i=1}^{n_{\text{train}}} |e_i|$, $L2$ error which is defined as $\sum_{i=1}^{n_{\text{train}}} e_i^2$, pseudo-$R^2$ which is defined as $1 - L2 / (\sum_{i=1}^{n_{\text{train}}} \squared{y_i - \bar{y}})$ and root mean squared error which is defined as $\sqrt{L2 / n_{\text{train}}}$. We also include random forests model creation times via the package \pkg{randomForest} \citep{Liaw2002} with its default settings.

We first note that Figure~\ref{fig:time_plots_large_sample} demonstrates that the \pkg{bartMachine} model creation runtime is approximately linear in $n$ (without in-sample statistics computed). There is about a 30\% speed-up when using four cores instead of one. The \textit{memcache} enhancement should be turned off only with sample sizes larger than $n=20,000$. Noteworthy is the 50\% reduction in time of constructing the model when not computing in-sample statistics. In-sample statistics are computed by default because the user generally wishes to see them. Also, for the purposes of this comparison, \pkg{BayesTree} models compute the in-sample statistics by necessity since the trees are not saved. The \pkg{randomForest} implementation becomes slower just after $n=1,000$ due to its reliance on a greedy exhaustive search at each node.

Figure~\ref{fig:time_plots_small_sample} displays results for smaller sample sizes ($n \leq 2,000$) that are often encountered in practice. We observe the \textit{memcache} enhancement provides about a 10\% speed improvement. Thus, if memory is an issue, it can be turned off with little performance degradation.

\subsection[Missing data in bartMachine]{Missing data in \pkg{bartMachine}}\label{subsec:missing_data}

\pkg{bartMachine} implements a native method for incorporating missing data into both model creation and future prediction with test data. The details are given in \citet{Kapelner2013} but we provide a brief summary here.

There are a number of ways to incorporate missingness into tree-based methods (see \citealp{Ding2010} for a review). The method implemented here is known as \qu{Missing Incorporated in Attributes} \citep[MIA, ][section 2]{Twala2008} which natively incorporates missingness by augmenting the nodes' splitting rules to (a) also handle sorting the missing data to the left or right and (b) use missingness \textit{itself} as a variable to be considered in a splitting rule. Table \ref{tab:mia} summarizes these new splitting rules as they are implemented within the package.

Implementing MIA into the BART procedure is straightforward. These new splitting rules are sampled uniformly during the GROW or CHANGE steps. For example, a splitting rule might be \qu{$\x_j < c$ or $\x_j$ is missing.} To account for splitting on missingness itself, we create dummy vectors of length $n$ for each of the $p$ attributes, denoted $\M_1, \ldots, \M_p$, which assume the value 1 when the entry is missing and 0 when the entry is present. The original training matrix is then augmented with these dummies, giving the opportunity to select missingness \textit{itself} when choosing a new splitting rule during the grow or change steps. Note that this can increase the number of predictors by up to a factor of 2. We illustrate building a \pkg{bartMachine} model with missing data in Section~\ref{subsec:incorporating_missing_data}. As described in \citet[][Section 6]{Chipman2010}, BART's runtime increases negligibly in the number of covariates and this has been our experience using the augmented training matrix.

\begin{table}[htp]
\caption{The MIA choices for all attributes $j \in \braces{1, \ldots, p}$ and all split points $x^*_{ij}$ where $i \in \braces{1,\ldots,n}$ during a GROW or CHANGE step in \pkg{bartMachine}.}
\begin{tabular}{ll}
\hline
1: & If $x_{ij}$ is missing, send it $\longleftarrow$; if it is present and $x_{ij} \leq x^*_{ij}$, send it $\longleftarrow$, otherwise $\longrightarrow$. \\
2: & If $x_{ij}$ is missing, send it $\longrightarrow$; if it is present and $x_{ij} \leq x^*_{ij}$, send it $\longleftarrow$, otherwise $\longrightarrow$. \\
3: & If $x_{ij}$ is missing, send it $\longleftarrow$; if it is present, send it $\longrightarrow$. \\
\hline
\end{tabular}
\label{tab:mia}
\end{table}

\subsection{Variable selection}\label{subsec:variable_selection}

Our package also implements the variable selection procedures developed in \citet{Bleich2014}, which is best applied to data problems where the number of covariates influencing the response is small relative to the total number of covariates. We give a brief summary of the procedures here. 

 In order to select variables, we make use of the \qu{variable inclusion proportions,} the proportion of times each predictor is chosen as a splitting rule divided by the total number of splitting rules appearing in the model (see Section~\ref{subsec:variable_importance} for more details). The variable selection procedure can be outlined as follows: 

\begin{enumerate}[1.]
\item Compute the model's variable inclusion proportions.
\item Permute the response vector, thereby breaking the relationship between the covariates and the response. Rebuild the model and compute the \qu{null} variable inclusion proportions. Repeat this a number of times to create a null permutation distribution. 
\item Three selection rules can be used depending on the desired stringency of selection:
\begin{enumerate}
\item Local Threshold: Include a predictor $\x_k$ if its variable inclusion proportion exceeds the $1-\alpha$ quantile of its own null distribution.
\item Global Max Threshold: Include a predictor $\x_k$ if its variable inclusion proportion exceeds the $1-\alpha$ quantile of the distribution of the maximum of the null variable inclusion proportions from each permutation of the response.
\item Global SE Threshold: Select $\x_k$ if its variable inclusion proportion exceeds a threshold based from its own null distribution mean and SD with a global multiplier shared by all predictors.
\end{enumerate}

\end{enumerate}

The Local procedure is the least stringent in terms of selection and the Global Max procedure the most. The Global SE procedure is a compromise, but behaves more similarly to the Global Max. \citet{Bleich2014} demonstrate that the best procedure depends on the underlying sparsity of the problem, which is often unknown. Therefore, the authors include an additional procedure that chooses the best of these thresholds via cross-validation and this method is also implemented in \pkg{bartMachine}.  As highlighted in \citet{Bleich2014}, this method performs favorably compared to variable selection using random forests \qu{importance scores}, which rely on the reduction in out-of-bag forecasting accuracy that occurs from shuffling the values for a particular predictor and dropping the out-of-bag observations down each tree.  Examples of these procedures for variable selection are provided in Section~\ref{subsec:variable_selection_regression}. 

\section{bartMachine Package Features for Regression}\label{sec:regression_features}

We illustrate the package features by using both real and simulated data, focusing first on regression problems. \pagebreak

\subsection{Computing parameters}\label{subsec:computing_parameters}

We first set some computing parameters. In this exploration, we allow up to 5GB of RAM for the Java heap\footnote{Note that the maximum amount of memory can be set only \textit{once} at the beginning of the \proglang{R} session (a limitation of \pkg{rJava} since only one Java Virtual Machine can be initiated per \proglang{R} session), but the number of cores can be respecified at any time.} and we set the number of computing cores available for use to 4.

\begin{Code}
R> options(java.parameters = "-Xmx5000m")
R> library("bartMachine")
R> set_bart_machine_num_cores(4)
\end{Code}

The following Sections~\ref{subsec:model_building} -- \ref{subsec:variable_selection_regression} use a dataset obtained from UCI \citep{Bache2013}. The $n=201$ observations are automobiles and the goal is to predict each automobile's price from 25 features (15 continuous and 10 nominal), first explored by \citet{Kibler1989}.\footnote{We first preprocess the data. We first drop one of the nominal predictors (car company) due to too many categories (22). We then coerce two of the of the nominal predictors to be continuous. Further, the response variable, price, was logged to reduce right skew in its distribution.} This dataset also contains missing data. The following code loads the data. We omit missing data for now and we create a variable for the design matrix $\X$ and the response $\y$ which has already been log-transformed.

\begin{Code}
R> data(automobile)
R> automobile = na.omit(automobile)
R> y <- automobile$log_price
R> X <- automobile; X$log_price <- NULL
\end{Code}

\subsection{Model building}\label{subsec:model_building}

We are now are ready to construct a \pkg{bartMachine} model. The default hyperparameters generally follow the recommendations of \citet{Chipman2010} and provide a ready-to-use algorithm for many data problems. Our hyperparameter settings are $m = 50$,\footnote{In contrast to \citet{Chipman2010}, we recommend this default as a good starting point rather than $m=200$ due to our experience experimenting with the \qu{RMSE by number of trees} feature found in later in this section. Performance is often similar and computational time and memory requirements are dramatically reduced.} $\alpha = 0.95$, $\beta = 2$, $k = 2$, $q = 0.9$, $\nu = 3$, and probabilities of the GROW / PRUNE / CHANGE steps is 28\% / 28\% /44\%. We set the number of burn-in Gibbs samples to 250 and number of post-burn-in samples to 1,000. We default the missing data feature to be off. We default the covariates to be equally important \textit{a priori}. Other parameters and their defaults can be found in the package's online manual. Below is a default \pkg{bartMachine} model. Here, $\X$ denotes automobile attributes and $\y$ denotes the log price of the automobile.

\begin{Code}
R> bart_machine <- bartMachine(X, y)
Building bartMachine for regression ... evaluating in sample data...done
\end{Code}

If one wishes to see more information about the individual iterations of the Gibbs sampler of Equation~\ref{eq:gibbs_sampler}, the flag \texttt{verbose} can be set to ``TRUE.'' One can see more debug information from the \proglang{Java} program by setting the flag \texttt{debug\_log} to true and the program will print to \texttt{unnamed.log} in the current working directory. In Figure~\ref{code:default_bart_summary} we inspect the model object to query its in-sample performance and to be reminded of the input data and model hyperparameters.

\begin{figure}[htp]
\centering
\begin{Code}
R> bart_machine
bartMachine v1.1.1 for regression

training data n = 160 and p = 41 
built in 1.7 secs on 4 cores, 50 trees, 250 burn-in and 1000 post. samples

sigsq est for y beforehand: 0.014 
avg sigsq estimate after burn-in: 0.00794 

in-sample statistics:
 L1 = 8.01 
 L2 = 0.65 
 rmse = 0.06 
 Pseudo-Rsq = 0.979
p-val for shapiro-wilk test of normality of residuals: 0.04584 
p-val for zero-mean noise: 0.97575
\end{Code}
\caption{The summary for the default \pkg{bartMachine} model built with the automobile data. Note that the \qu{p-val for shapiro-wilk test of normality of residuals} is marginally less than 5\%. Thus we conclude that the noise of Equation~\ref{eq:bart} is not normally distributed. Just as when interpreting the results from a linear model, non-normality implies we should be circumspect concerning \pkg{bartMachine} output that relies on this distributional assumption such as the credible and prediction intervals of Section~\ref{subsec:credible_and_prediction_intervals}.}
\label{code:default_bart_summary}
\end{figure}

Since the response was considered continuous, we employ \pkg{bartMachine} for regression. The dimensions of the design matrix are given. Note that we dropped 41 observations that contained missing data (which we will retain in Section~\ref{subsec:incorporating_missing_data}).  We then have a recording of the MSE for the OLS model and our average estimate of $\sigsq_e$. We are then given in-sample statistics on error. Pseudo-$R^2$ is calculated via $1 - SSE/SST$. Also provided are outputs from tests of the error distribution being mean centered and normal. In this case, we cannot conclude normality of the residuals using the Shapiro-Wilk test.

We can also obtain out-of-sample statistics to assess level of overfitting by using k-fold cross-validation. Using 10 randomized folds we find:

\begin{Code}
R> k_fold_cv(X, y, k_folds = 10)
..........
$L1_err
[1] 21.64303

$L2_err
[1] 4.742511

$rmse
[1] 0.1721647

$PseudoRsq
[1] 0.8467881
\end{Code}

The Pseudo-$R^2$ being lower out-of-sample versus in-sample suggests evidence that \pkg{bartMachine} is slightly overfitting (note also that the training sample during cross-validation is 10\% smaller). This function also returns the $\yhat$ predictions as well as the vector of the fold indices (which are omitted above). 

It may also be of interest how the number of trees $m$ affects performance. One can examine how out-of-sample predictions vary by the number of trees via 

\begin{figure}[htp]
\centering
\includegraphics[width=3.5in]{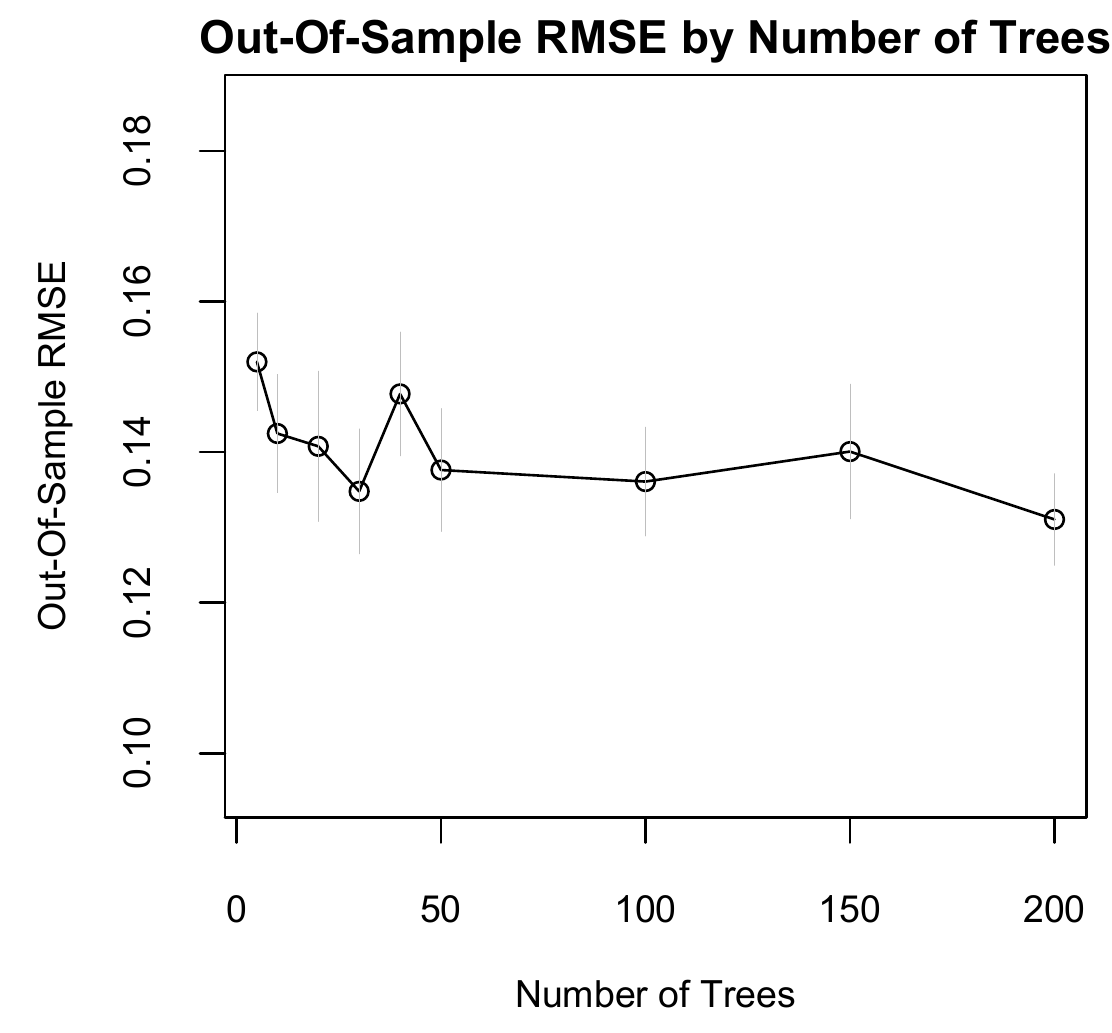}
\caption{Out-of-sample predictive performance by number of trees}
\label{fig:rmse_by_trees}
\end{figure}

\begin{Code}
R> rmse_by_num_trees(bart_machine, num_replicates = 20)
\end{Code}

\noindent and the output is shown in Figure~\ref{fig:rmse_by_trees}.

It seems that increasing $m > 50$ does not result in any substantial increase in performance. We can now try to build a better \pkg{bartMachine} by grid-searching over a set of hyperparameter combinations, including $m$ \citep[for more details, see BART-cv in][]{Chipman2010}. The default grid search is small and it can be customized by the user.

\begin{Code}
R> bart_machine_cv <- bartMachineCV(X, y)
...
bartMachine CV win: k: 2 nu, q: 3, 0.9 m: 200
\end{Code}

This function returns the \qu{winning} model, which is the one with lowest out-of-sample RMSE over a 5-fold cross-validation. Here, the cross-validated \pkg{bartMachine} model has slightly better in-sample performance (L1 = 8.18, L2 = 0.68 and Pseudo-$R^2 = 0.978$) as well as slightly better out-of-sample performance (L1 = 21.05, L2 = 4.40 and Pseudo-$R^2 = 0.858$) which were evaluated via:

\begin{Code}
R> k_fold_cv(X, y, k_folds = 10, k = 2, nu = 3, q = 0.9, num_trees = 200)
\end{Code}

\noindent Predictions are handled with the \texttt{predict} function. Fits for the first seven rows are:

\begin{Code}
R> predict(bart_machine_cv, X[1 : 7, ])
[1]  9.494941  9.780791  9.795532 10.058445  9.670211  9.702682  9.911394
\end{Code}

\noindent We also include a convenience method \texttt{bart\_predict\_for\_test\_data} that will predict and return out-of-sample error metrics when the test outcomes are known.

\subsection{Assumption checking}\label{subsec:assumption_checking}

The package includes features that assess the plausibility of the BART model assumptions. Checking the mean-centeredness of the noise is addressed in the summary output of Figure~\ref{code:default_bart_summary} and is simply a one-sample $t$-test of the average residual value against a null hypothesis of true mean zero. We assess both normality and heteroskedasticity via:

\begin{Code}
R> check_bart_error_assumptions(bart_machine_cv)
\end{Code}

\begin{figure}[h]
\centering
\includegraphics[width=5in]{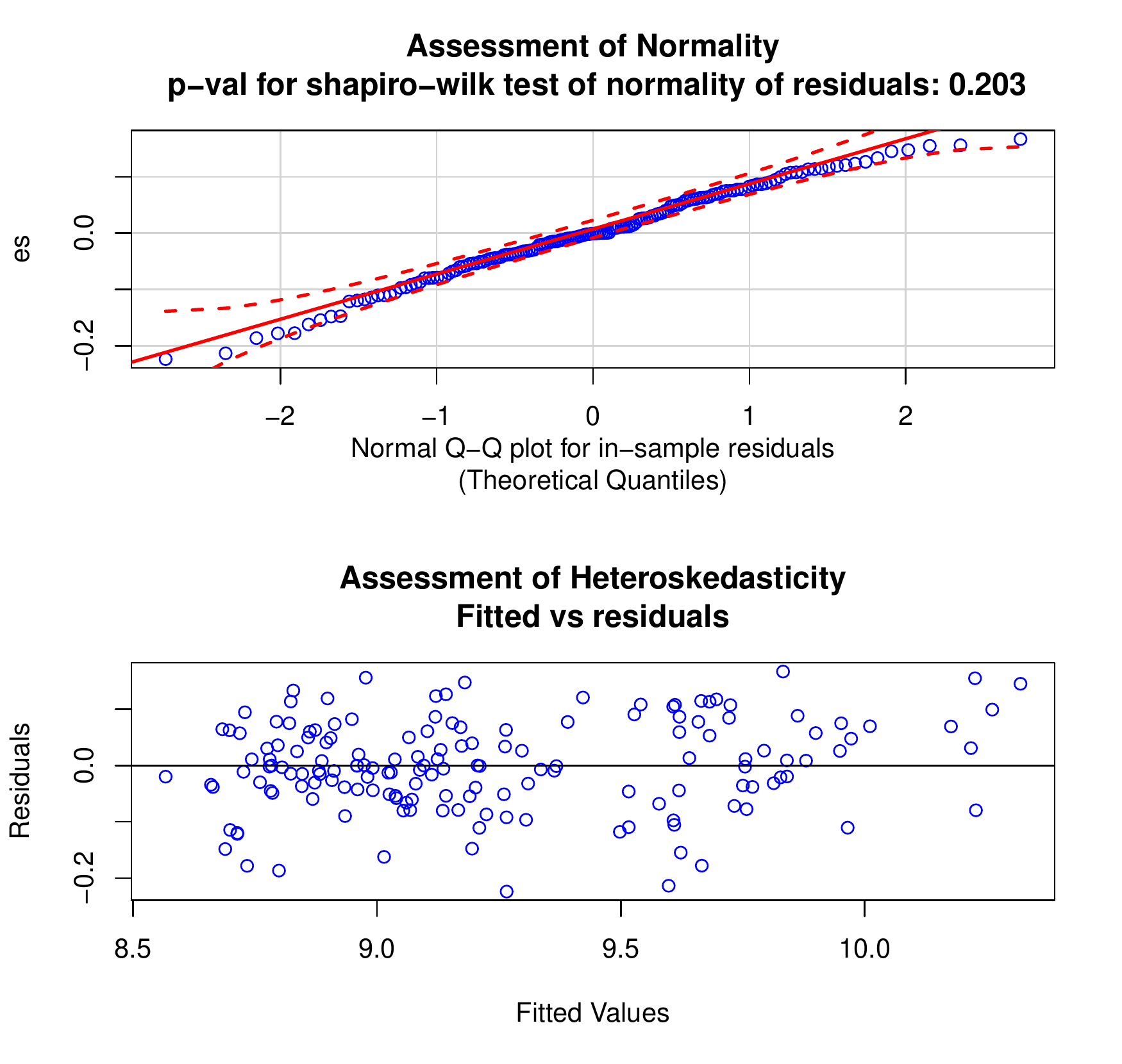}
\caption{Test of normality of errors using QQ-plot and the Shapiro-Wilk test (top), residual plot to assess heteroskedasticity (bottom).}
\label{fig:bart_normality_heteroskedasticity}
\end{figure}

This will display a plot similar to Figure~\ref{fig:bart_normality_heteroskedasticity} which contains a QQ-plot (to assess normality) as well as a residual-by-predicted plot (to assess homoskedasticity). It appears that the errors are most likely normal and homoskedastic.

In addition to the model assumptions, BART requires convergence of its Gibbs sampler which can be investigated via:

\begin{Code}
R> plot_convergence_diagnostics(bart_machine_cv)
\end{Code}

\begin{figure}[htp]
\centering
\includegraphics[width=6.1in]{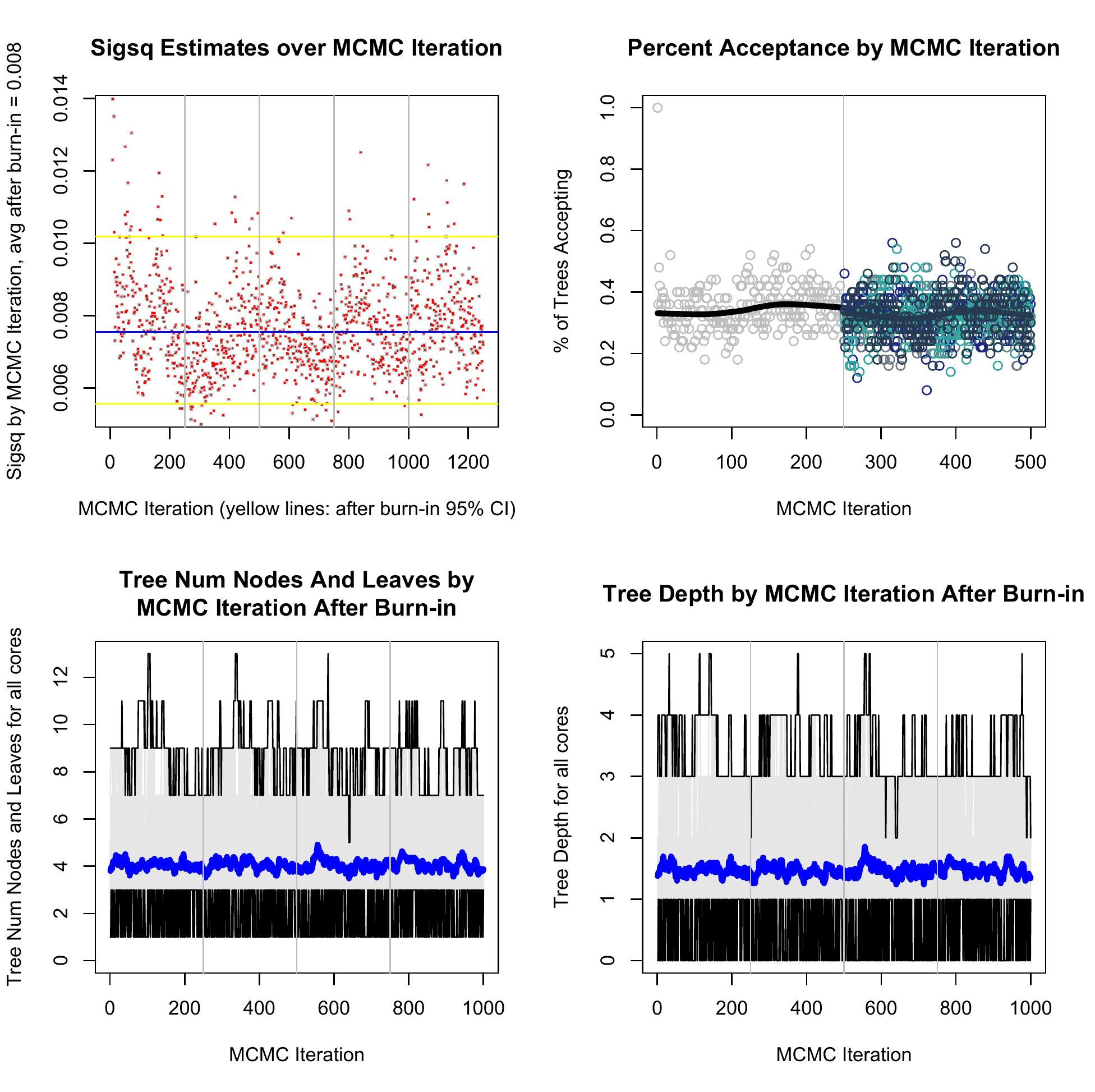}
\caption{Convergence diagnostics for the cross-validated \pkg{bartMachine} model. Top left: $\sigsq$ by MCMC iteration. Samples to the left of the first vertical grey line are burn-in from the first computing core's MCMC chain. The four subsequent plots separated by grey lines are the post-burn-in iterations from each of the four computing cores employed during model construction. Top right: percent acceptance of Metropolis-Hastings proposals across the $m$ trees where each point plots one iteration. Points before the grey vertical line illustrate burn-in iterations and points after illustrate post burn-in iterations. Each computing core is colored differently. Bottom left: average number of leaves across the $m$ trees by iteration (post burn-in only where computing cores separated by vertical grey lines). Bottom right: average tree depth across the $m$ trees by iteration (post burn-in only where computing cores separated by vertical grey lines).}
\label{fig:convergence_diagnostics}
\end{figure}

Figure~\ref{fig:convergence_diagnostics} displays the plot which features four types of convergence diagnostics (each are detailed in the figure caption). It appears that the \pkg{bartMachine} model has been sufficiently burned-in.

\subsection{Credible intervals and prediction intervals}\label{subsec:credible_and_prediction_intervals}

An advantage of BART is that if we believe the priors and model assumptions, the Bayesian probability model and corresponding burned-in MCMC iterations provide the approximate posterior distribution of $f\parens{\x}$. Thus, one can compute uncertainty estimates via quantiles of the posterior samples. These provide Bayesian \qu{credible intervals} which are intervals for the conditional expectation function, $\cexpe{\y}{\X}$.

Another useful uncertainty interval can be computed for individual predictions by combining uncertainty from the conditional expectation function with the systematic, homoskedastic normal noise produced by $\errorrv$. This is accomplished by generating 1,000 samples (by defauilt) from the posterior predictive distribution and then reporting the appropriate quantiles.

Below is an example of how both types of intervals are computed in the package (for the 100th observation of the training data):

\begin{Code}
R> calc_credible_intervals(bart_machine_cv, X[100, ], ci_conf = 0.95)
     ci_lower_bd ci_upper_bd
[1,]    8.725202    8.971687
R> calc_prediction_intervals(bart_machine_cv, X[100, ], pi_conf = 0.95)
     pi_lower_bd pi_upper_bd
[1,]    8.631243     9.06353
\end{Code}
 
Note that the prediction intervals are wider than the credible intervals because they reflect the uncertainty from the error term.

We can then plot these intervals in sample:

\begin{Code}
R> plot_y_vs_yhat(bart_machine_cv, credible_intervals = TRUE)
R> plot_y_vs_yhat(bart_machine_cv, prediction_intervals = TRUE)
\end{Code}

\begin{figure}[htp]
\centering
\begin{subfigure}[c]{\textwidth}
                \centering
                \includegraphics[width=3.5in]{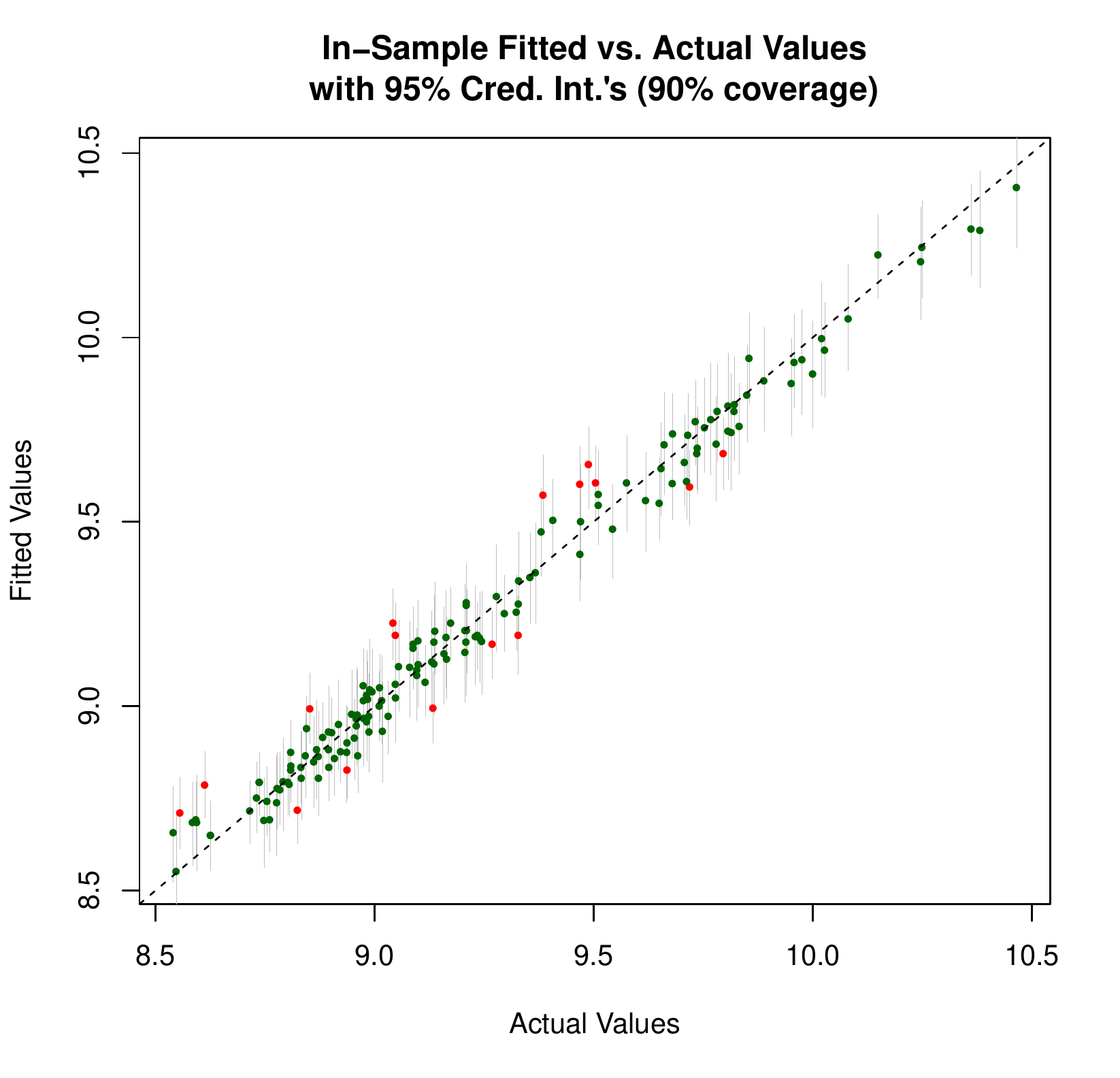}
                \caption{Segments illlustrate credible intervals}
                \label{fig:plot_y_vs_y_hat_with_credible_intervals}
        \end{subfigure}\\
\begin{subfigure}[c]{\textwidth}
                \centering
                \includegraphics[width=3.5in]{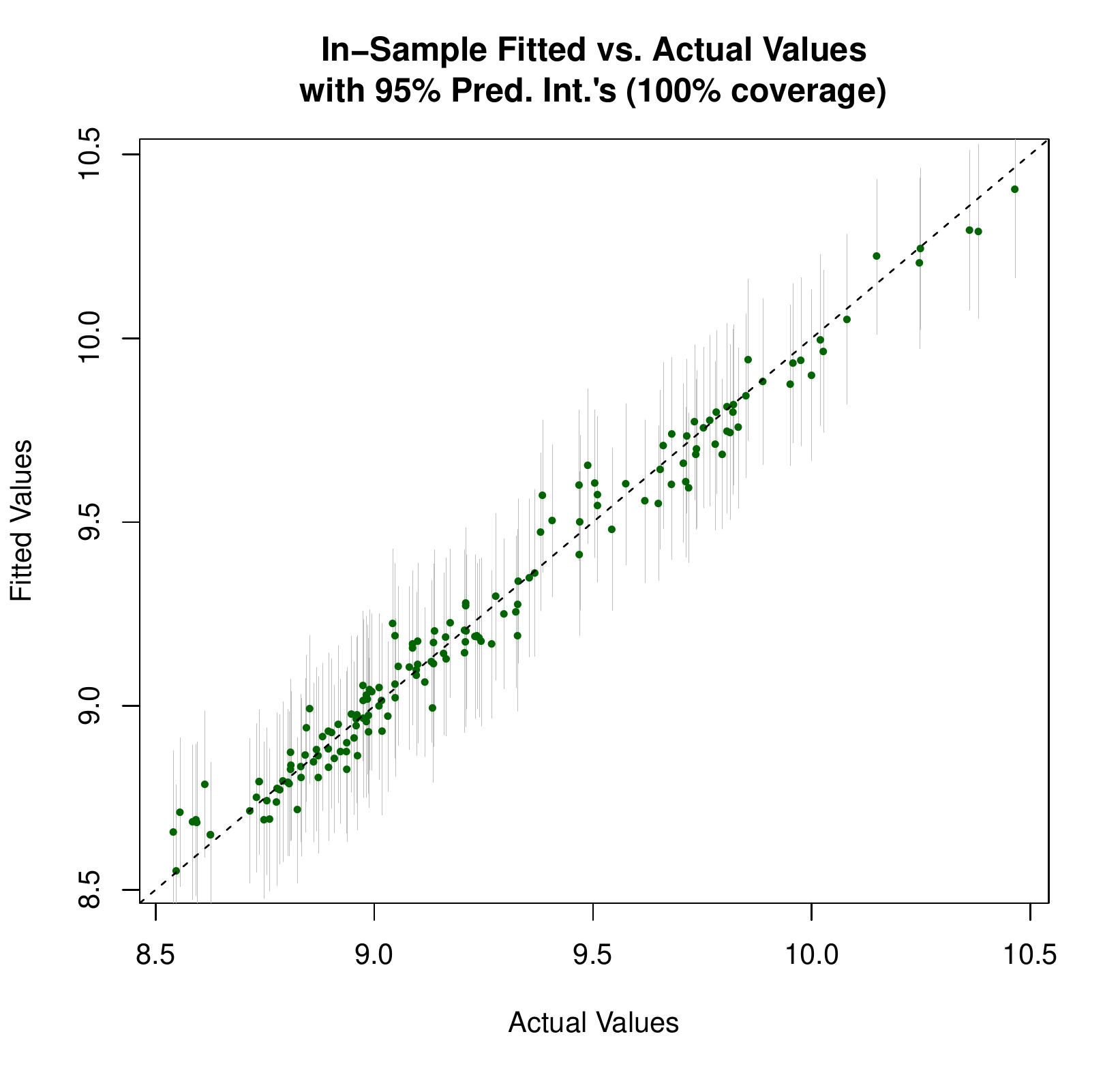}
                \caption{Segments illlustrate prediction intervals}
                \label{fig:plot_y_vs_y_hat_with_prediction_intervals}
        \end{subfigure}
\caption{Fitted versus actual response values for the automobile dataset. Segments are 95\% credible intervals (a) or 95\% prediction intervals (b). Green dots indicate the true response is within the stated interval and red dots indicate otherwise. Note that the percent coverage in (a) is not expected to be 95\% because the response includes a noise term.}
\label{fig:plot_y_vs_y_hat}
\end{figure}

Figure~\ref{fig:plot_y_vs_y_hat_with_credible_intervals} shows how our prediction fared against the original response (in-sample) with 95\% credible intervals.  Figure~\ref{fig:plot_y_vs_y_hat_with_prediction_intervals} shows the same prediction versus the original response plot now with 95\% prediction intervals.

\subsection{Variable importance}\label{subsec:variable_importance}

\begin{figure}[htp]
\centering
\includegraphics[width=5.9in]{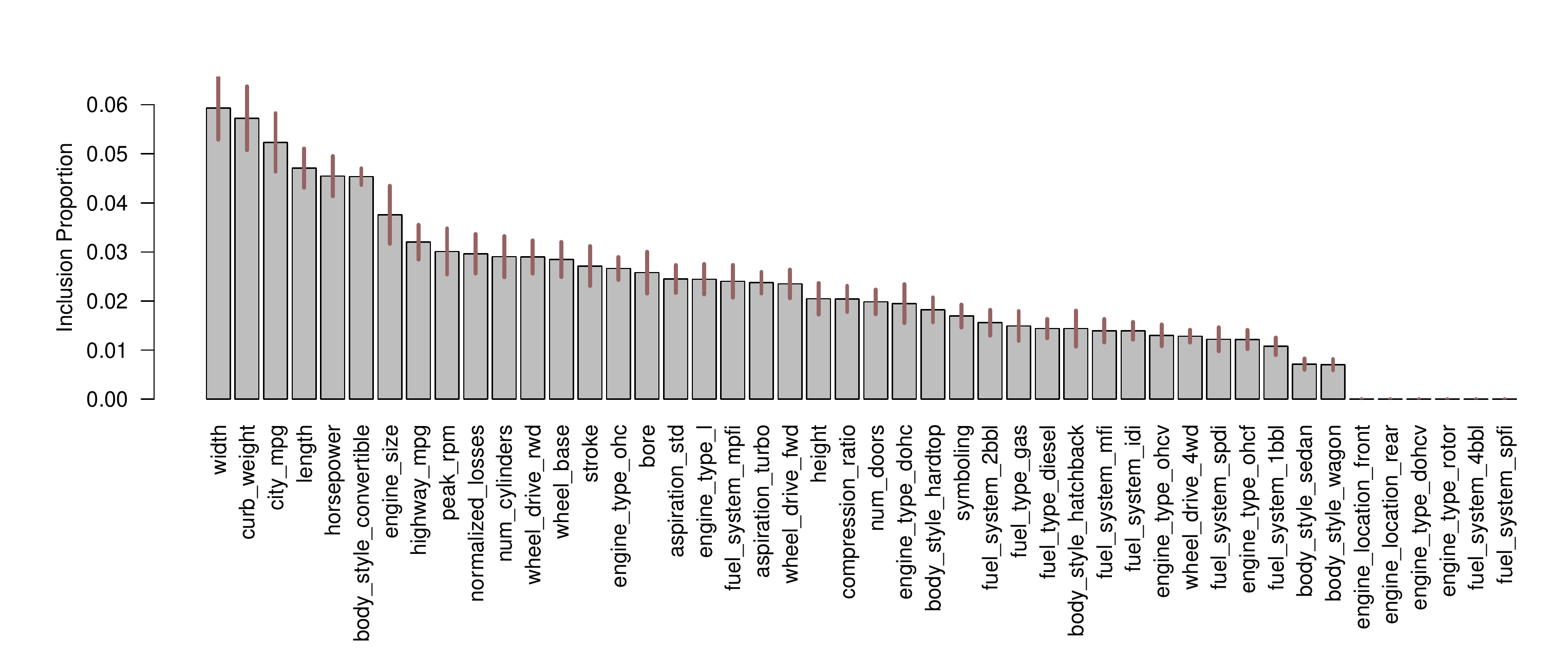}
\caption{Average variable inclusion proportions in the cross-validated \pkg{bartMachine} model for the automobile data averaged over 100 model constructions to obtain stable estimates across many posterior modes in the sum-of-trees distribution (as recommended in \citealp{Bleich2014}). The segments atop the bars represent 95\% confidence intervals. The eight predictors with inclusion proportions of zero feature identically one value (after missing data was dropped).}
\label{fig:var_imp_automobile_cc}
\end{figure}

After a \pkg{bartMachine} model is built, it is natural to ask the question: which variables are most important? This is assessed by examining the splitting rules in the $m$ trees across the post burn-in MCMC iterations which are known as \qu{inclusion proportions} \citep{Chipman2010}. The inclusion proportion for any given predictor represents the proportion of times that variable is chosen as a spliting rule out of all splitting rules among the posterior draws of the sum-of-trees model. Figure~\ref{fig:var_imp_automobile_cc} illustrates the inclusion proportions for all variables obtained via:

\begin{Code}
R> investigate_var_importance(bart_machine_cv, num_replicates_for_avg = 20)
\end{Code}

Selection of variables which \textit{significantly} affect the response is addressed briefly in Section~\ref{subsec:variable_selection}, examples are provided in Section~\ref{subsec:variable_selection_regression} but for full treatment of this feature, please see \citet{Bleich2014}. 

\subsection{Variable effects}\label{subsec:variable_effects}

It is also natural to ask: does $\x_j$ affect the response, controlling for other variables in the model? This is roughly analogous to the $t$-test in ordinary least squares regression of no linear effect of $\x_j$ on $\y$ while controlling for $\x_{-j}$. The null hypothesis here is the same but the linearity constraint is relaxed. To test this, we employ a permutation approach where we record the observed Pseudo-$R^2$ from the \pkg{bartMachine} model built with the original data. Then we permute the $\x_j$th column, thereby destroying any relationship between $\x_j$ and $\y$, construct a new duplicate \pkg{bartMachine} model from this permuted design matrix and record a ``null'' Pseudo-$R^2$. We then repeat this  process to obtain a null distribution of Pseudo-$R^2$'s. Since the alternative hypothesis is that $\x_j$ has an effect on $\y$ in terms of predictive power, our $p$ value is the proportion of null Pseudo-$R^2$'s greater than the observed Pseudo-$R^2$, making our procedure a natural one-sided test. Note, however, that this test is conditional on the BART model and its selected priors being true, similar to the assumptions of the linear model. 

If we wish to test if a set of covariates $A \subset \braces{\x_{1}, \ldots, \x_{p}}$ affect the response after controlling for other variables, we repeat the procedure outlined in the above paragraph by permuting the columns of $A$ in every null sample. We do not permute each column separately, but instead permute as a unit in order to perserve collinearity structure. This is roughly analogous to the partial $F$-test in ordinary least squares regression. 

If we wish to test if \textit{any} of the covariates matter in predicting $\y$, we simply permute $\y$ during the null sampling. This procedure breaks the relationship between the response and the predictors but does not alter the existing associations between predictors. This is roughly analogous to the omnibus $F$-test in ordinary least squares regression.

\begin{figure}[h]
\centering
\begin{subfigure}[c]{0.48\textwidth}
                \centering
                \includegraphics[width=3.2in]{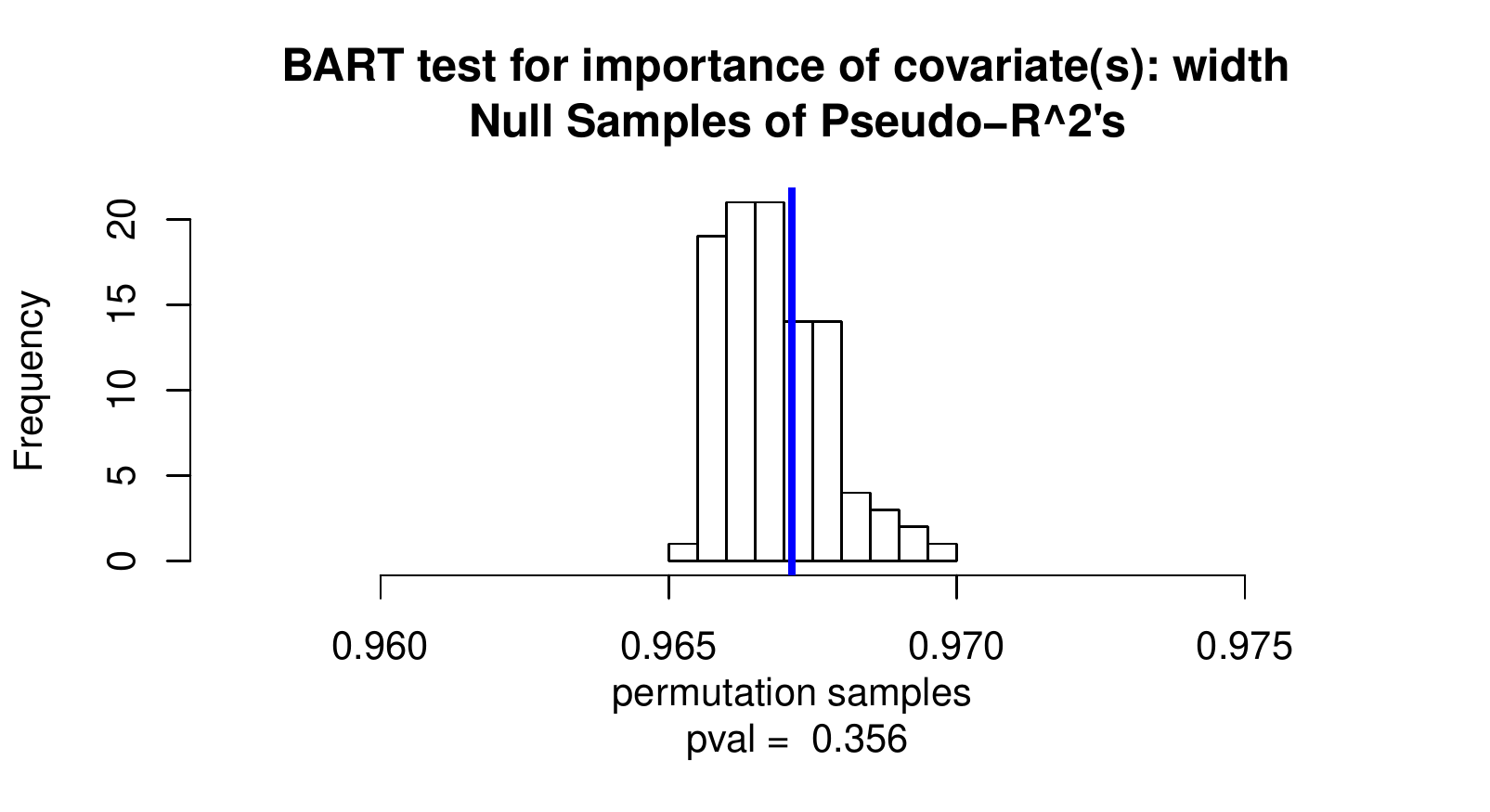}
                \caption{\texttt{width}}
                \label{fig:cov_test_width}
        \end{subfigure}~~
\begin{subfigure}[c]{0.48\textwidth}
                \centering
                \includegraphics[width=3.2in]{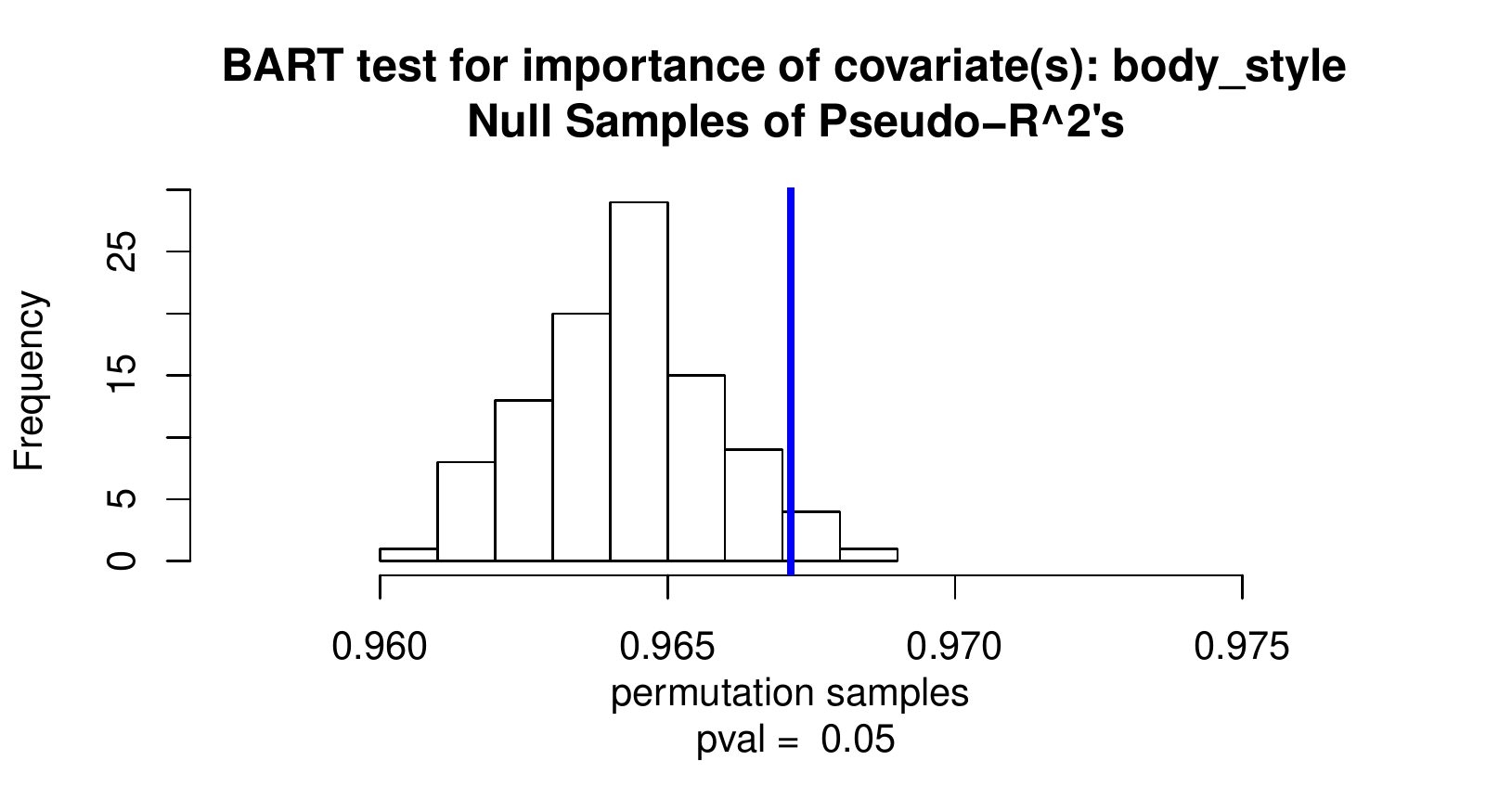}
                \caption{\texttt{body style}}
                \label{fig:cov_test_body_style}
        \end{subfigure}\\
\begin{subfigure}[b]{0.48\textwidth}
                \centering
                \includegraphics[width=3.2in]{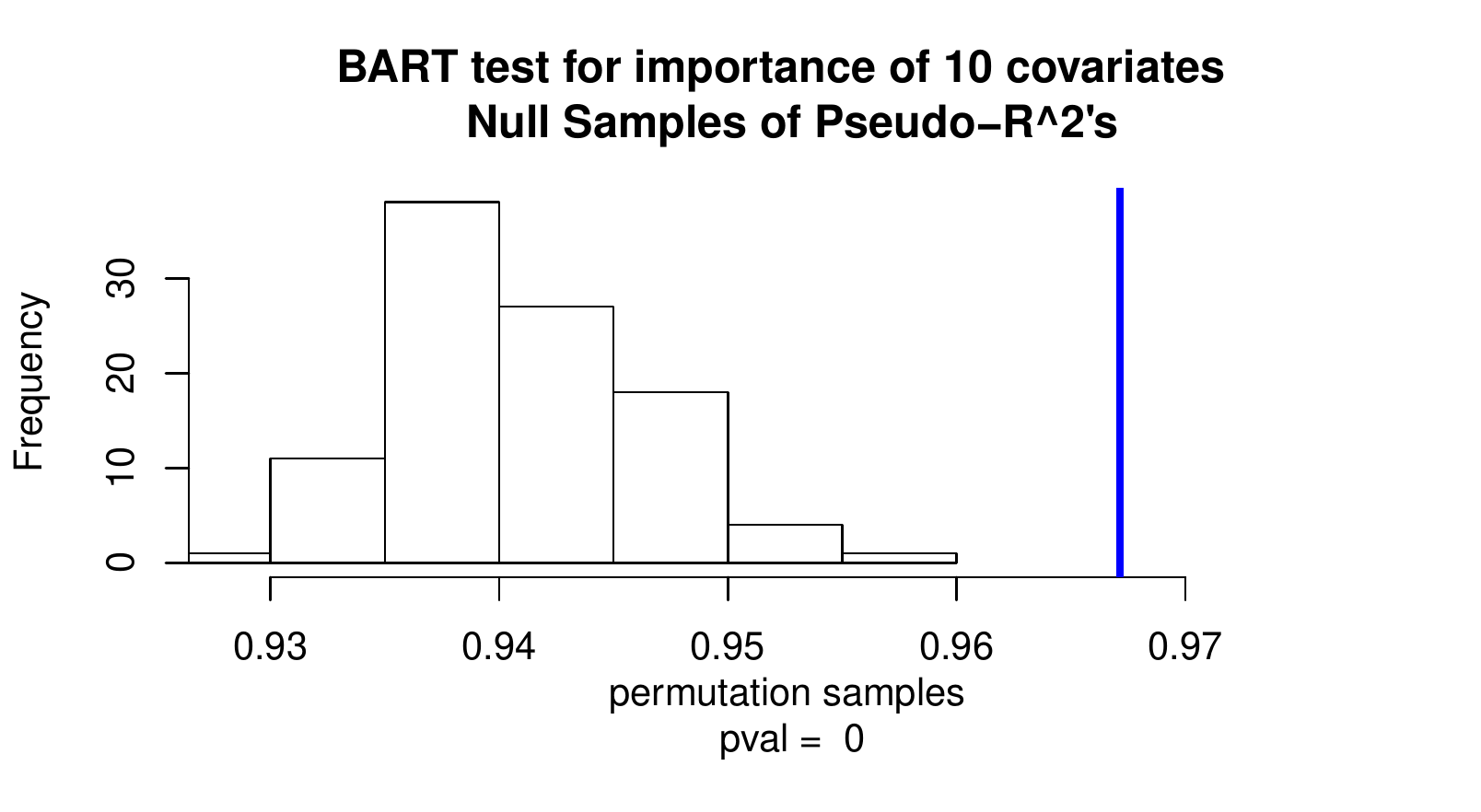}
                \caption{The 10 most split-on variables}
                \label{fig:cov_test_top_10}
        \end{subfigure}~~
\begin{subfigure}[b]{0.48\textwidth}
                \centering
                \includegraphics[width=3.2in]{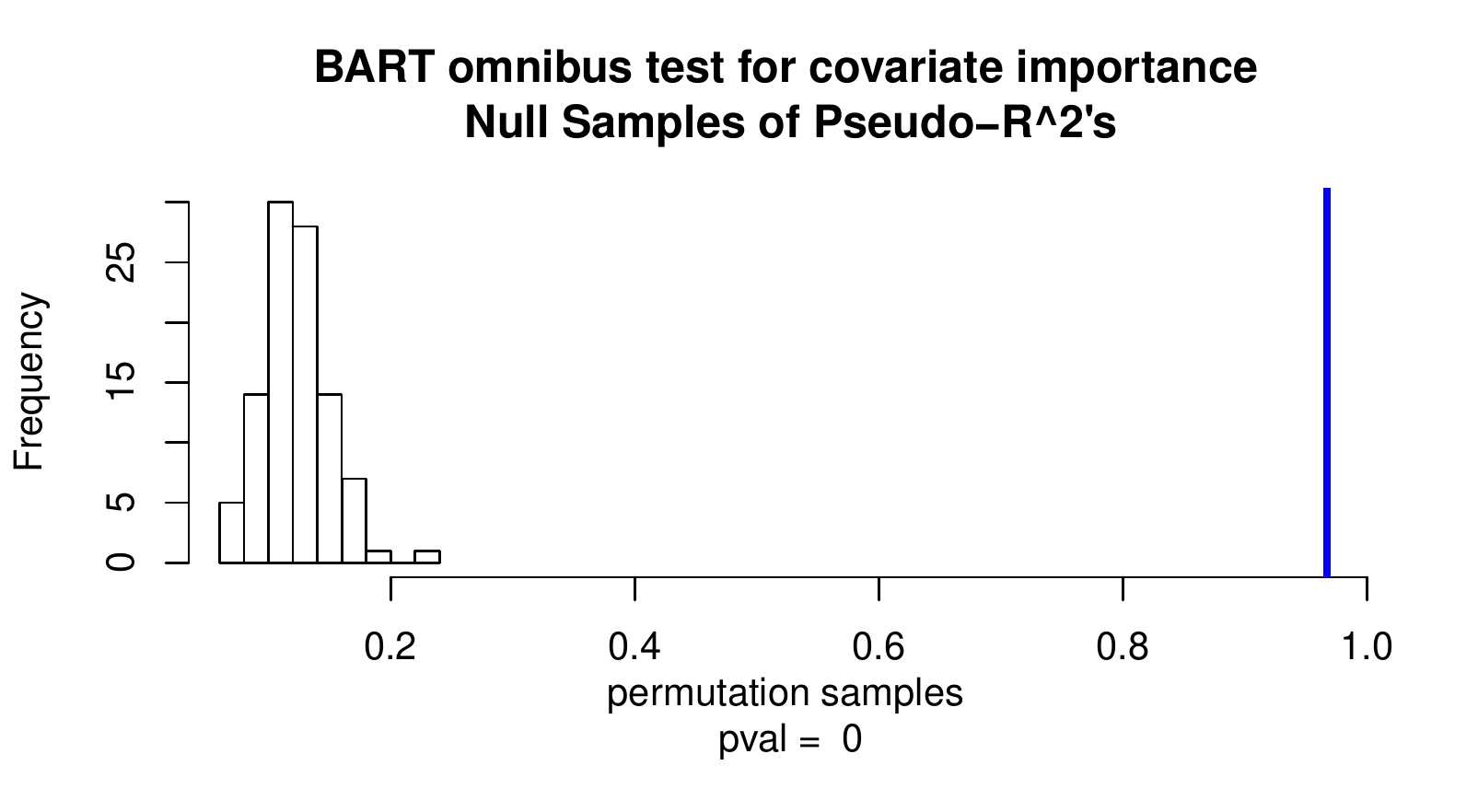}
                \caption{All covariates}
                \label{fig:cov_test_omnibus}
        \end{subfigure}
\caption{Tests of covariate importance conditional on the cross-validated \pkg{bartMachine} model. All tests performed with 100 null samples.}
\label{fig:cov_tests}
\end{figure}

At $\alpha = 0.05$, Figure~\ref{fig:cov_test_width} demonstrates an insignificant effect of the variable \texttt{width} of car on price. Even though \texttt{width} is putatively the \qu{most important} variable as measured by proportions of splits in the posterior sum-of-trees model (Figure~\ref{fig:var_imp_automobile_cc}), note that this is largely an easy prediction problem with many collinear predictors. Figure~\ref{fig:cov_test_body_style} shows the results of a test of the putatively most important categorical variable, \texttt{body style} (which involves permuting the categories, then dummifying the levels to preserve the structure of the variable). We find a marginally significant effect ($p = 0.0495$). A test of the top ten most important variables is convincingly significant (Figure~\ref{fig:cov_test_top_10}). For the omnibus test, Figure~\ref{fig:cov_test_omnibus} illustrates an extremely statistically significant result, as would be expected. The code to run these tests is shown below (output suppressed).

\begin{Code}
R> cov_importance_test(bart_machine_cv, covariates = c("width"))
R> cov_importance_test(bart_machine_cv, covariates = c("body_style"))
R> cov_importance_test(bart_machine_cv, covariates = c("width",
  "curb_weight", "city_mpg", "length", "horsepower", "body_style", 
  "engine_size", "highway_mpg", "peak_rpm", "normalized_losses"))
R> cov_importance_test(bart_machine_cv)
\end{Code}

\subsection{Partial dependence}\label{subsec:partial_dependence}

A data analyst may also be interested in understanding how $\x_j$ affects the response on average, after controlling for other predictors. This can be examined using \citet{Friedman2001}'s Partial Dependence Function (PDP),

\bneqn\label{eq:true_pdp}
f_j(\x_j) = \expesub{\x_{-j}}{f(\x_j,\x_{-j})} = \int f(\x_j,\x_{-j}) \mathrm{dP}\parens{\x_{-j}},
\eneqn

\noindent where $\x_{-j}$ denotes all variables except $\x_j$. The PDP of predictor $\x_j$ gives the average value of $f$ when $\x_j$ is fixed and $\x_{-j}$ varies over its marginal distribution, $\mathrm{dP}\parens{\x_{-j}}$. As neither the true model $f$ nor the distribution of the predictors $\mathrm{dP}\parens{\x_{-j}}$ are known, we estimate Equation~\ref{eq:true_pdp} by computing

\begin{figure}[htp]
\centering
\begin{subfigure}[c]{0.48\textwidth}
                \centering
                \includegraphics[width=3.2in]{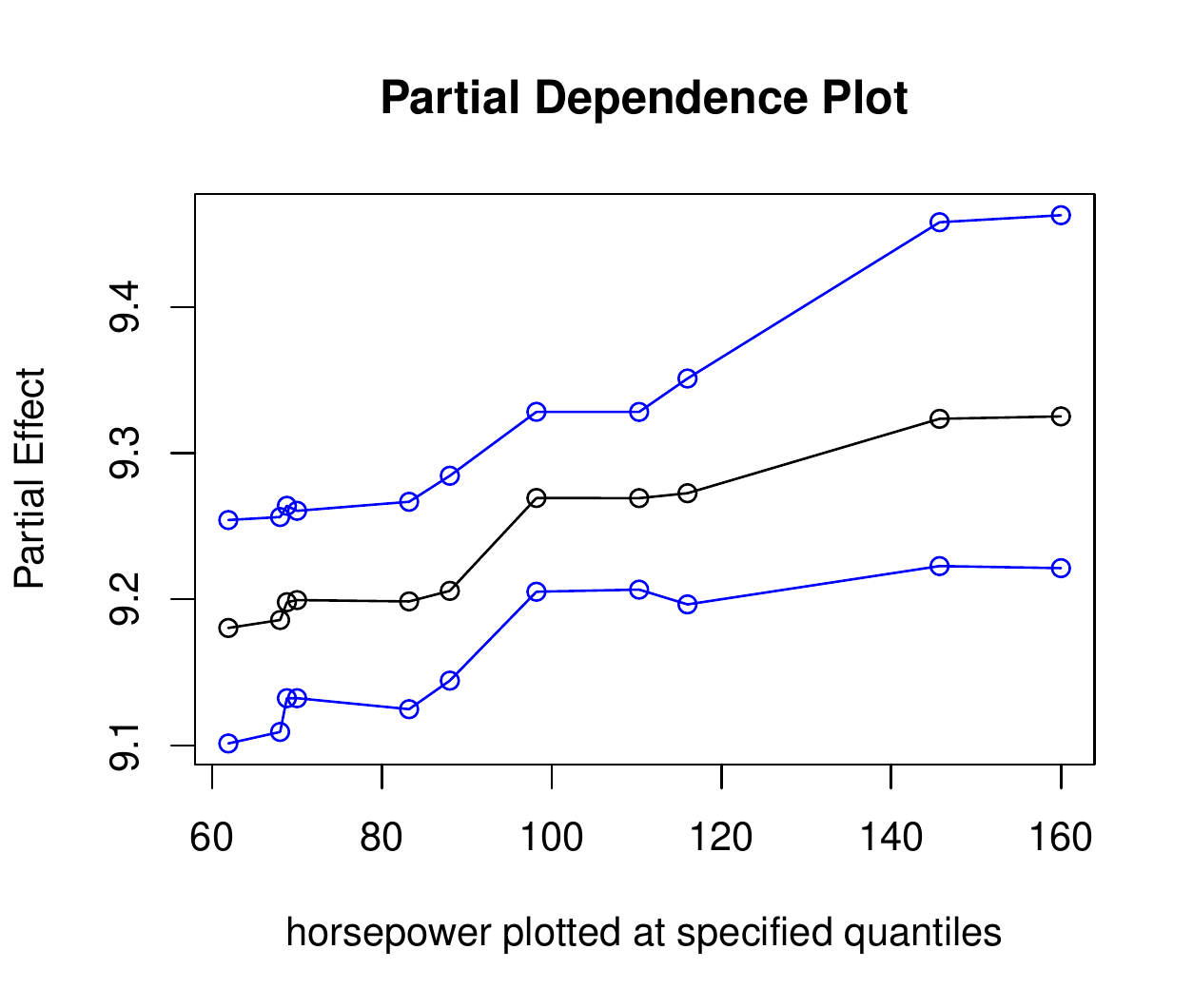}
                \caption{\texttt{horsepower}}
                \label{fig:pdp_horsepower}
        \end{subfigure}~~
\begin{subfigure}[c]{0.48\textwidth}
                \centering
                \includegraphics[width=3.2in]{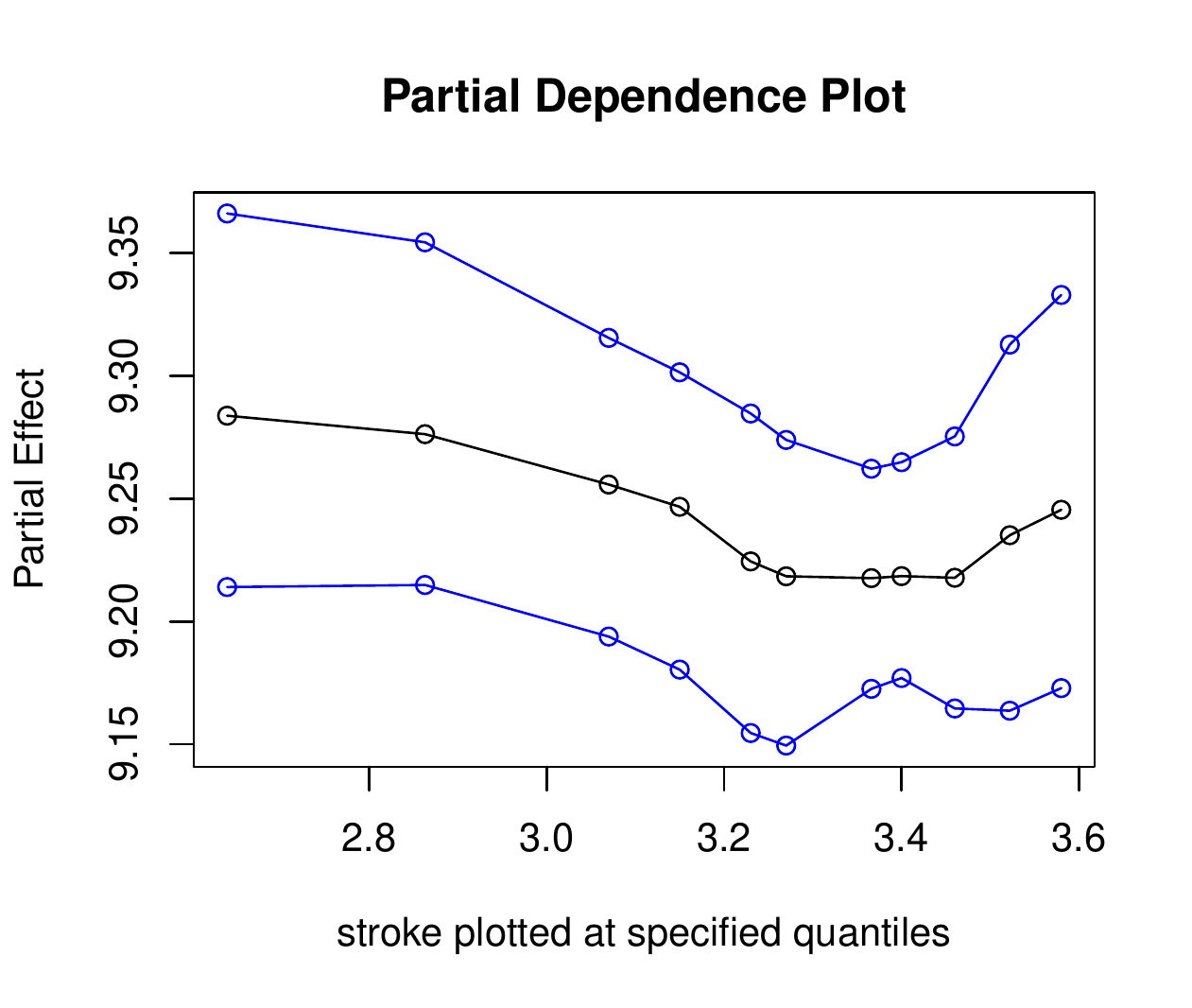}
                \caption{\texttt{stroke}}
                \label{fig:pdp_stroke}
        \end{subfigure}
\caption{PDPs plotted in black and 95\% credible intervals plotted in blue for variables in the automobile dataset. Points plotted are at the 5\%ile, 10\%ile, 20\%ile, \ldots, 90\%ile and 95\%ile of the values of the predictor. Lines plotted between the points approximate the PDP by linear interpolation.}
\label{fig:pdps}
\end{figure}

\bneqn\label{eq:est_pdp}
\hat{f}_j(\x_j) = \frac{1}{n}\sum\limits_{i=1}^n \hat{f}(\x_{j},\x_{-j,i})
\eneqn

\noindent where $n$ is the number of observations in the training data and $\hat{f}$ denotes the \pkg{bartMachine} model. Since BART provides an estimated posterior distribution, we can plot credible bands for the PDP function. In Equation~\ref{eq:est_pdp}, the $\hat{f}$ can be replaced with a function that calculates the $q$th quantile of the post-burned-in MCMC iterations for $\yhat$. Figure~\ref{fig:pdp_horsepower} plots the PDP along with the 2.5\%ile and the 97.5\%ile for the variable \texttt{horsepower}. By varying over most of the range of \texttt{horsepower}, the price is predicted to increase by about \$1000. Figure~\ref{fig:pdp_stroke} plots the PDP along with the 2.5\%ile and the 97.5\%ile for the variable \texttt{stroke}. This predictor seemed to be relatively unimportant according to Figure~\ref{fig:var_imp_automobile_cc} and the PDP confirms this, with a very small, yet nonlinear average partial effect. The code for both plots is below.

\begin{Code}
R> pd_plot(bart_machine_cv, j = "horsepower")
R> pd_plot(bart_machine_cv, j = "stroke")
\end{Code}

\subsection{Incorporating missing data}\label{subsec:incorporating_missing_data}

The procedure for incorporating missing data was introduced in Section~\ref{subsec:missing_data}. We now build a \pkg{bartMachine} model using this procedure below:

\begin{Code}
R> y <- automobile$log_price
R> X <- automobile; X$log_price <- NULL
R> bart_machine <- bartMachine(X, y, use_missing_data = TRUE, 
      use_missing_data_dummies_as_covars = TRUE)
R> bart_machine
bartMachine v1.1.1 for regression

Missing data feature ON
training data n = 201 and p = 50 
built in 1.4 secs on 1 core, 50 trees, 250 burn-in and 1000 post. samples

sigsq est for y beforehand: 0.016 
avg sigsq estimate after burn-in: 0.00939 

in-sample statistics:
 L1 = 11.49 
 L2 = 1.04 
 rmse = 0.07 
 Pseudo-Rsq = 0.9794
p-val for shapiro-wilk test of normality of residuals: 0.69814 
p-val for zero-mean noise: 0.96389 
\end{Code}

Note that we now use the complete data set including the 41 observations for which there were missing features. Also note that $p$ has now increased from 41 to 50. The nine \qu{new} predictors are:

\begin{Code}
[1] "engine_location_rear" "engine_type_rotor"    "fuel_system_4bbl"    
[4] "fuel_system_spfi"     "M_normalized_losses"  "M_bore"              
[7] "M_stroke"             "M_horsepower"         "M_peak_rpm"
\end{Code}

The first two predictors are two new levels for the variable \texttt{engine\_location} which appear in the 41 rows with missingness. The next two predictors are two new levels for the variable \texttt{fuel\_system} which appear in the 41 rows with missingness as well. The last five new predictors are dummy variables which indicate missingness constructed from the predictors which exhibited missingness (due to the \texttt{use\_missing\_data\_dummies\_as\_covars} parameter being set to true).

The procedure of Section~\ref{subsec:missing_data} also natively incorporates missing data during prediction. Missingness will yield larger credible intervals. In the example below, we suppose that the \texttt{curb\_weight} and \texttt{symboling} values were suddenly unavailable for the 20th automobile and we observe their credible intervals widening as a result.

\begin{Code}
R> x_star <- X[20, ]
R> calc_credible_intervals(bart_machine, x_star, ci_conf = 0.95)
     ci_lower_bd ci_upper_bd
[1,]    8.650093    8.824515
R> x_star[c("curb_weight", "symboling")] <- NA
R> calc_credible_intervals(bart_machine, x_star, ci_conf = 0.95)
     ci_lower_bd ci_upper_bd
[1,]    8.622582    8.978313
\end{Code}

\subsection{Variable selection}\label{subsec:variable_selection_regression}

\begin{figure}[htp]
\centering
\includegraphics[width=5.6in]{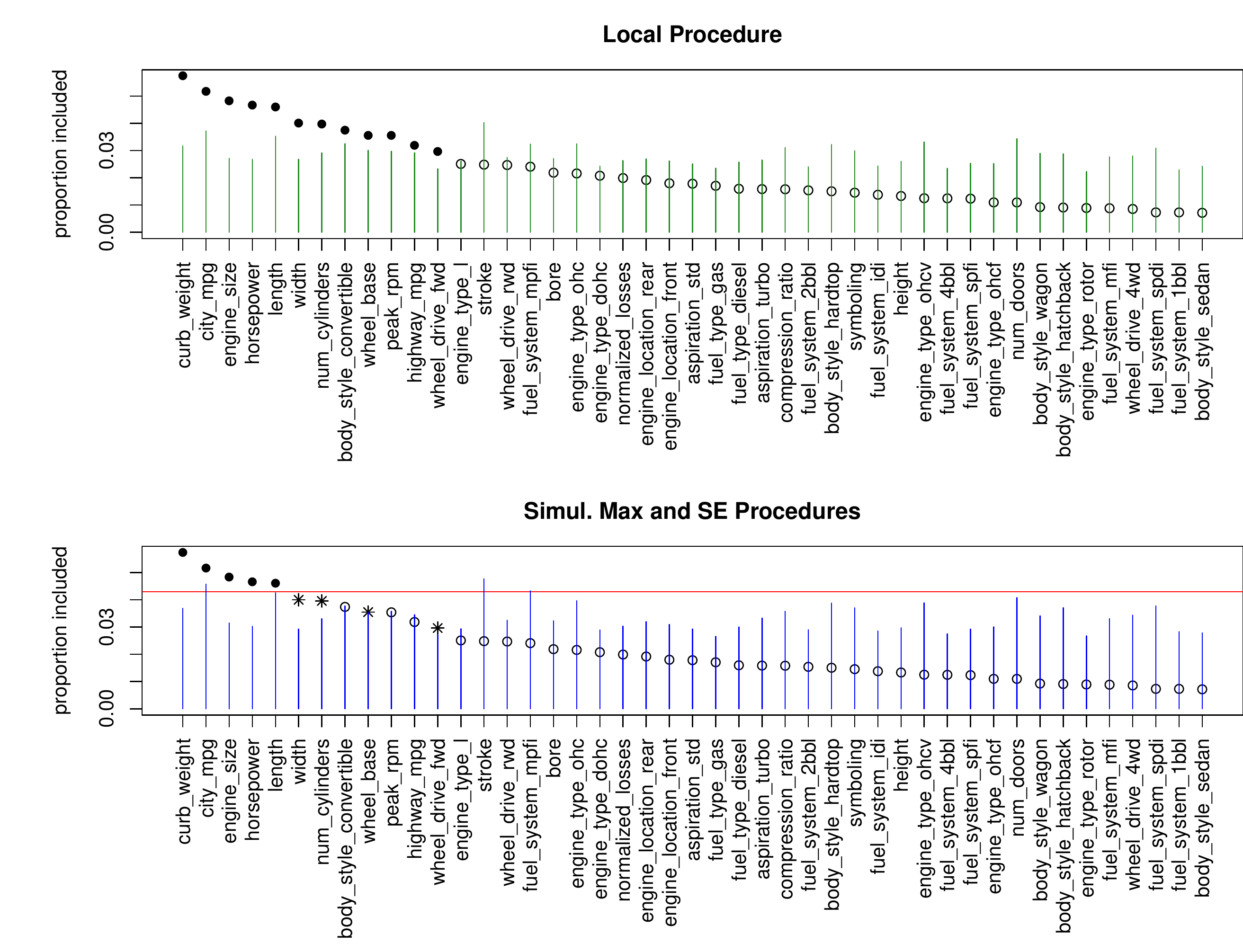}
\caption{Visualization of the three variable selection procedures outlined in Section~\ref{subsec:variable_selection} with $\alpha = 0.05$. The top plot illustrates the ``Local'' procedure. The green lines are the threshold levels determined from the permutation distributions that must be exceeded for a variable to be selected. The plotted points are the variable inclusion proportions for the observed data (averaged over five duplicate \pkg{bartMachine} models). If the observed value is higher than the green bar, the variable is included and is displayed as a solid dot; if not, it is not included and it is displayed as an open dot. The bottom plot illustrates both the ``Global SE'' and ``Global Max'' thresholds. The red line is the cutoff for ``Global Max'' and variables pass this threshold are displayed as solid dots. The blue lines represent the thresholds for the ``Global SE'' procedure. Variables that exceed this cutoff but not the ``Global Max'' threshold are displayed as asterisks. Open dots exceed neither threshold.}
\label{fig:var_selection_plot}
\end{figure}

In this section we demonstrate the principled variable selection procedure introduced in Section~\ref{subsec:variable_selection} and developed in detail in \citet{Bleich2014}. The following code will select variables based on the three thresholds and also displays the plot in Figure~\ref{fig:var_selection_plot}.\footnote{By default, variable selection is performed individually on dummy variables for a factor. The variable selection procedures return the permutation distribution and an aggregation of the dummy variables' inclusion proportions can allow for variable selection to be performed on an entire factor.}

\begin{Code}
R> vs <- var_selection_by_permute(bart_machine, 
            bottom_margin = 10, num_permute_samples = 10)
R> vs$important_vars_local_names
  "curb_weight"  "city_mpg" "engine_size"   "horsepower"            
  "length"       "width"    "num_cylinders" "body_style_convertible"
  "wheel_base"   "peak_rpm" "highway_mpg"   "wheel_drive_fwd"       
R> vs$important_vars_global_max_names
  "curb_weight" "city_mpg"    "engine_size" "horsepower"  "length"     
R> vs$important_vars_global_se_names
  "curb_weight"   "city_mpg"      "engine_size" "horsepower" "length"           
  "width"         "num_cylinders" "wheel_base"  "wheel_drive_fwd"
\end{Code}

Usually,  ``Global Max'' and ``Global SE'' perform similarly, as they are both more stringent in selection. However, in many situations it will not be clear to the data analyst which threshold is most appropriate. The ``best'' procedure can be chosen via cross-validation on out-of-sample RMSE as follows: 

\begin{Code}
var_selection_by_permute_response_cv(bart_machine)
$best_method
[1] "important_vars_local_names"

$important_vars_cv
 [1] "body_style_convertible" "city_mpg"               "curb_weight"           
 [4] "engine_size"            "engine_type_ohc"        "horsepower"            
 [7] "length"                 "num_cylinders"          "peak_rpm"              
[10] "wheel_base"             "wheel_drive_fwd"        "wheel_drive_rwd"       
[13] "width" 
\end{Code}

On this dataset, the \qu{best} approach (as defined by out-of-sample prediction error) is the ``Local'' procedure.\\

\subsection{Informed prior information on covariates}\label{subsec:cov_prior}

\citet{Bleich2014} propose a method for incorporating informed prior information about the predictors into the BART model. This can be achieved by modifying the prior on the splitting rules as well as the corresponding calculations in the Metropolis-Hastings step. In particular, covariates believed to influence the response can be proposed more often as candidates for splitting rules. Useful prior information can aid in both variable selection and prediction tasks. We illustrate the impact of a correctly informed prior in the context of the \citet{Friedman1991} function (Equation~\ref{eq:friedman}). 

\bneqn\label{eq:friedman}
\y = 10\sin{\pi\x_1\x_2}+20(\x_3-.5)^2+10\x_4+5\x_5 + \berrorrv, \qquad \berrorrv \sim \multnormnot{n}{\0}{\sigsq\bv I}.
\eneqn

To illustrate, we construct a design matrix $\X$ where the first five columns are predictors which influence the response ($\x_1, \ldots, \x_5$ in Equation~\ref{eq:friedman}) and the next 95 columns are predictors that do not affect the response.

All that is required is a specification of relative weights for each predictor. These are converted internally to probabilities. We assign 5 times the weight to the 5 true covariates of the model relative to the 95 useless covariates.

\begin{Code}
R> prior <- c(rep(5, times = 5), rep(1, times = 95))
\end{Code}

We now sample 500 observations from the Friedman function and construct a default \pkg{bartMachine} model as well as a \pkg{bartMachine} model with the informed prior and compare their performance on a test set of another 500 observations.

\begin{Code}
R> bart_machine <- bartMachine(X, y)
R> bart_machine_informed <- bartMachine(X, y, cov_prior_vec = prior)

R> bart_predict_for_test_data(bart_machine, Xtest, ytest)$rmse
[1] 1.661159
R> bart_predict_for_test_data(bart_machine_informed, Xtest, ytest)$rmse
[1] 1.232925
\end{Code}

There is a substantial improvement in out-of-sample predictive performance when a properly informed prior is used. 

Note that by default the prior vector down-weights the indicator variables that result from dummifying factors so that the total set of dummy variables has the same weight as a continuous covariate.

\subsection{Interaction effect detection}\label{subsec:interaction_detection}

In Section~\ref{subsec:variable_importance}, we explored using variable inclusion proportions to understand the relative influences of different covariates. A similar procedure can be carried out for examining interaction effects within a BART model. This question was initially explored in \citet{Damien2013} where an interaction was considered to exist between two variables if they both appeared in at least one splitting rule in a given tree. We refine the definition of an interaction as follows. 

We first begin with a $p \times p$ matrix of zeroes. Within a given tree, for each split rule variable $j$, we look at all split rule variables of child nodes, $k$, and we increment the $j, k$ element of the matrix. Hence variables are considered to interact in a given tree \textit{only if} they appear together in a contiguous downward path from the root node to a terminal node.  Note that a variable may interact with itself (when fitting a linear effect, for instance). Since there is no order between the parent and child, we then add the $j, k$ counts together with the $k, j$ counts (if $j \neq k$). Summing across trees and MCMC iterations gives the total number of interactions for each pair of variables from which relative importance can be assessed. 

We demonstrate interaction detection on the Friedman function using 10 covariates using the code below: 

\begin{Code}
R> interaction_investigator(bart_machine, num_replicates_for_avg = 25, 
      num_var_plot = 10, bottom_margin = 5)
\end{Code}

\begin{figure}[h]
\centering
\includegraphics[width=4.5in]{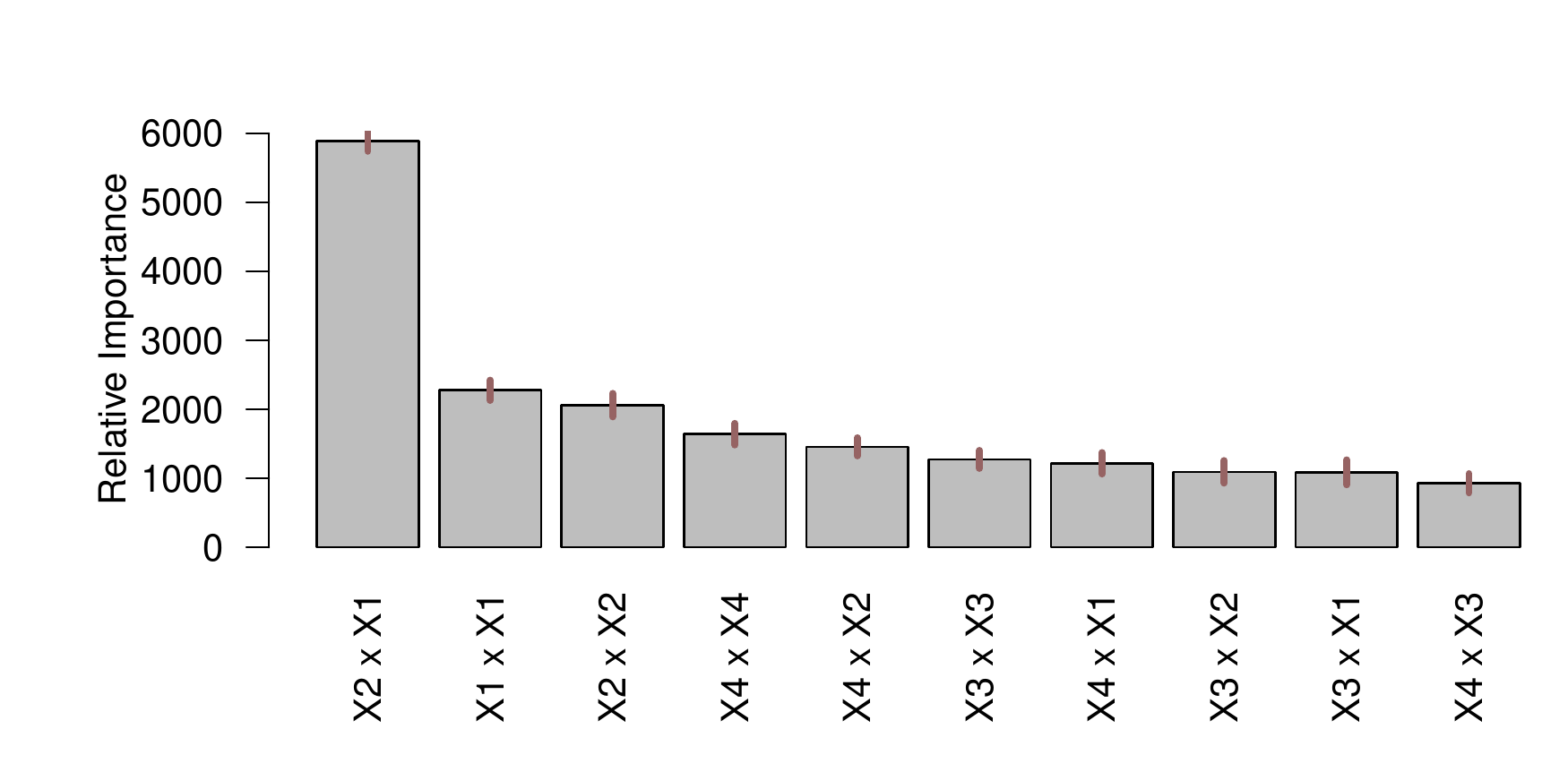}
\caption{The top 10 average variable interaction counts (termed \qu{relative importance}) in the default \pkg{bartMachine} model for the Friedman function data averaged over 25 model constructions. The segments atop the bars represent 95\% confidence intervals.}
\label{fig:friedman_function_interactions}
\end{figure}

Shown in Figure~\ref{fig:friedman_function_interactions} are the ten most important interactions in the model. The illustration is averaged over many model constructions to obtain stable estimates across many posterior modes in the sum-of-trees distribution. Notice that the interaction between $\x_1$ and $\x_2$ dominates all other terms, as \pkg{bartMachine} is correctly capturing the single true interaction effect in Equation~\ref{eq:friedman}. Choosing which of these interactions \textit{significantly} affect the response is not addressed in this paper. The methods suggested in Section~\ref{subsec:variable_selection} may be applicable here and we consider this to be fruitful future work.

\subsection{bartMachine Model Persistence Across R Sessions}\label{subsec:persistence}

A convenient feature of \pkg{bartMachine} is its ability to \qu{serialize.} Serialization allows the user to construct models and have them persist across \code{R} sessions. The cost is time during model creation and hard drive space. Thus, the serialize feature is defaulted to \qu{off.} We demonstrate using the code below:

\begin{Code}
R> bart_machine <- bartMachine(X, y, serialize = TRUE)
R> save.image("bart_demo.RData")
R> q("no")
> R
R> options(java.parameters = "-Xmx2500m")
R> library(bartMachine)
R> load("bart_demo.RData")
R> predict(bart_machine, X[1 : 4, ])
[1] 20.0954617 14.8860727 10.9483889 11.4350277
\end{Code}

The training data is the same as in the previous section: $n=100$ and $p=10$. For the default \pkg{bartMachine} settings, $m=50$, number of burn-in MCMC iterations is 250 and number of posterior samples is 1000. These settings yielded an almost instant serialization, generating a 2.1MB RData file. For a more realistic dataset with $n=5000$, $p=1000$, $m=100$ and 5000 posterior samples, the serialization and saving of the file took a few minutes and requires 100MB.

Note that the package throws an error if the user attempts to make use of a \pkg{bartMachine} object in another session which was not serialized:

\begin{Code}
R> bart_machine <- bartMachine(X, y)
R> save.image("bart_demo.RData")
R> q("no")
> R
R> options(java.parameters = "-Xmx2500m")
R> library(bartMachine)
R> load("bart_demo.RData")
R> predict(bart_machine, X[1 : 4, ])
Error in check_serialization(object) : 
  This bartMachine object was loaded from an R image but was not serialized.
  Please build bartMachine using the option "serialize = TRUE" next time.
\end{Code}

\section{bartMachine Package Features for Classification}\label{sec:classification_features}

In this section we highlight the features that differ from the regression case when the response is dichotomous. The illustrative dataset consists of 332 Pima Indians obtained from the UCI repository. Of the 332 subjects, 109 were diagnosed with diabetes, the binary response variable. There are seven continuous predictors which are body metrics such as blood pressure, glucose concentration, etc. and there is no missing data. 
	
Building a \pkg{bartMachine} model for classification has the same computing parameters except that $q, \nu$ cannot be specified since there is no longer a prior on $\sigsq$ (see Section~\ref{subsec:probit_bart}). We first build a cross-validated model below.

\begin{Code}
R> bart_machine_cv <- bartMachineCV(X, y)
... bartMachine CV win: k: 3 m: 50
R> bart_machine_cv
Bart Machine v1.0b for classification

training data n = 332 and p = 7 
built in 0.5 secs on 4 cores, 50 trees, 250 burn-in and 1000 post. samples

confusion matrix:

           predicted No predicted Yes model errors
actual No       211.000         12.00        0.054
actual Yes       41.000         68.00        0.376
use errors        0.163          0.15        0.160
\end{Code}

Classification models have an added hyperparameter, \texttt{prob\_rule\_class}, which is the rule for determining if the probability estimate is great enough to be classified into the positive category. We can see above that the \pkg{bartMachine} at times predicts ``NO'' for true ``YES'' outcomes and we suffer from a 37.6\% error rate for this outcome. We can try to mitigate this error by lowering the threshold to increase the number of ``YES'' labels predicted:\\

\begin{Code}
R> bartMachine(X, y, prob_rule_class = 0.3)
Bart Machine v1.0b for classification

training data n = 332 and p = 7 
built in 0.5 secs on 4 cores, 50 trees, 250 burn-in and 1000 post. samples

confusion matrix:

           predicted No predicted Yes model errors
actual No       178.000        45.000        0.202
actual Yes       12.000        97.000        0.110
use errors        0.063         0.317        0.172
\end{Code}

This lowers the model error to 11\% for the ``YES'' class, but at the expense of increasing the error rate for the ``NO'' class. We encourage the user to cross-validate this rule based on the appropriate objective function for the problem at hand.

We can also check out-of-sample statistics:

\begin{Code}
R> oos_stats = k_fold_cv(X, y, k_folds = 10)
R> oos_stats$confusion_matrix
           predicted No predicted Yes model errors
actual No       203.000        20.000        0.090
actual Yes       47.000        62.000        0.431
use errors        0.188         0.244        0.202
\end{Code}

\noindent Note that it is possible to predict both class labels and probability estimates for given observations:

\begin{Code}
R> predict(bart_machine_cv, X[1 : 2, ], type = "prob")
[1] 0.6253160 0.1055975
R> predict(bart_machine_cv, X[1 : 2, ], type = "class")
[1] Yes No 
Levels: No Yes
\end{Code}

\begin{figure}[htp]
\centering
\includegraphics[width=4.2in]{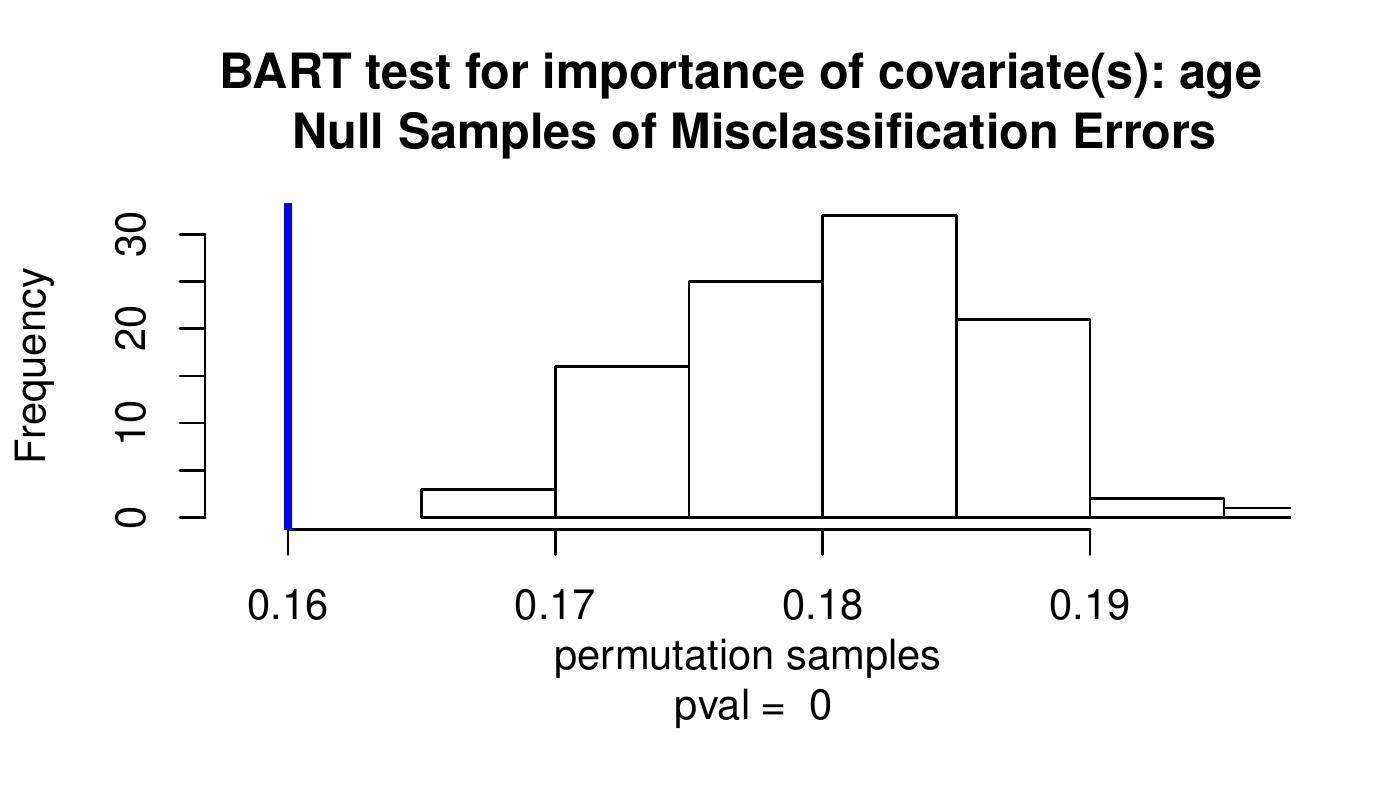}
\caption{Test of covariate importance for predictor \texttt{age} on whether or not the subject will contract \texttt{Diabetes}.}
\label{fig:covariate_test_age}
\end{figure}

When using the covariate tests of Section~\ref{subsec:variable_effects}, total misclassification error becomes the statistic of interest instead of Pseudo-$R^2$. The $p$ value is calculated now as the proportion of null samples with \textit{lower} misclassification error. Figure~\ref{fig:covariate_test_age} illustrates the test showing that predictor \texttt{age} seems to matter in the prediction of \texttt{Diabetes}, controlling for other predictors.

\begin{figure}[htp]
\centering
\includegraphics[width=3.3in]{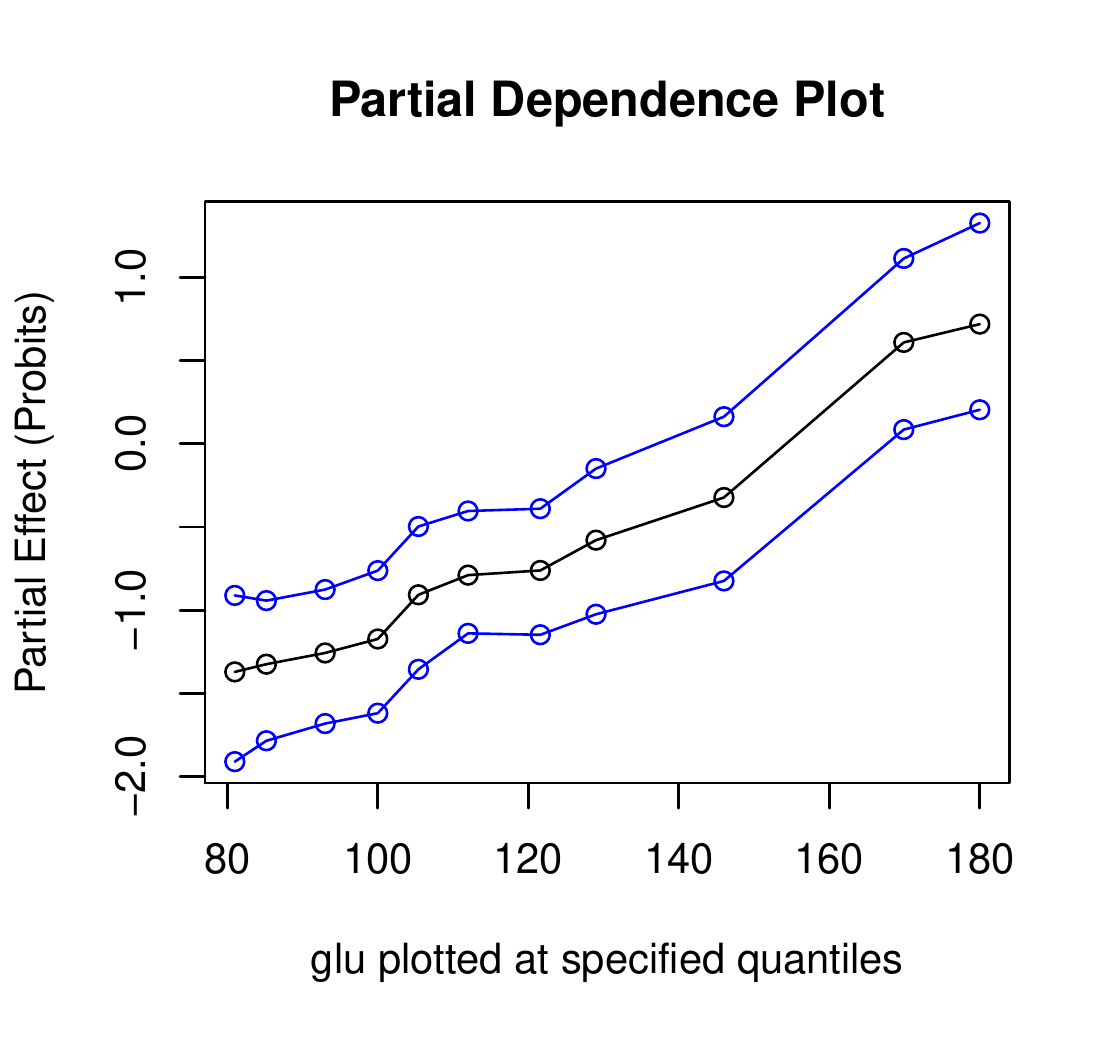}
\caption{PDP for predictor \texttt{glu}. The blue lines are 95\% credible intervals.}
\label{fig:glucose_partial_dependence}
\end{figure}

The partial dependence plots of Section~\ref{subsec:partial_dependence} are now scaled as probit of the probability estimate. Figure~\ref{fig:glucose_partial_dependence} illustrates that as glucose increases, the probability of contracting \texttt{Diabetes} increases linearly on a probit scale.

Credible intervals are implemented for classification \pkg{bartMachine} and are displayed on the probit scale. Note that the prediction intervals of Section~\ref{subsec:credible_and_prediction_intervals} do not exist for classification.

\begin{Code}
R> calc_credible_intervals(bart_machine_cv, X[1 : 2, ])
     ci_lower_bd ci_upper_bd
[1,]  0.34865355   0.8406097
[2,]  0.01686486   0.2673171
\end{Code}

Other functions work similarly to regression except those that plot the responses and those that explicitly depend on RMSE as an error metric.

\section{Discussion}\label{sec:discussion}

This article introduced \pkg{bartMachine}, a new \proglang{R} package which implements Bayesian additive regression trees. The goal of this package is to provide a fast, extensive and user-friendly implementation accessible to a wide range of data analysts, and increase the visibility of BART to a broader statistical audience. We hope we have provided organized, well-documented open-source code and we encourage the community to make innovations on this package.

\subsection*{Replication}

The stable version of \pkg{bartMachine} is on CRAN and the development version is located at \url{http://github.com/kapelner/bartMachine}. The package code is under the GPL3 and MIT licenses. Results, tables, and figures found in this paper can be replicated via the scripts located in the \texttt{bart\_package\_paper} folder within the \texttt{git} repository.

\subsection*{Acknowledgements}

We thank Richard Berk, Andreas Buja, Zachary Cohen, Ed George, Alex Goldstein, Shane Jensen, Abba Krieger, and Robert DeRubeis for helpful discussions. We thank Simon Urbanek for his very generous help with \pkg{rJava} at many stages of this research. Adam Kapelner acknowledges support from the National Science Foundation's Graduate Research Fellowship Program.

\bibliography{bart_package_paper}

\begin{thebibliography}{27}
\newcommand{\enquote}[1]{``#1''}
\providecommand{\natexlab}[1]{#1}
\providecommand{\url}[1]{\texttt{#1}}
\providecommand{\urlprefix}{URL }
\expandafter\ifx\csname urlstyle\endcsname\relax
  \providecommand{\doi}[1]{doi:\discretionary{}{}{}#1}\else
  \providecommand{\doi}{doi:\discretionary{}{}{}\begingroup
  \urlstyle{rm}\Url}\fi
\providecommand{\eprint}[2][]{\url{#2}}

\bibitem[{Albert and Chib(1993)}]{Albert1993}
Albert J, Chib S (1993).
\newblock \enquote{{Bayesian Analysis of Binary and Polychotomous Response
  Data}.}
\newblock \emph{Journal of the American Statistical Association},
  \textbf{88}(422), 669--679.

\bibitem[{Bache and Lichman(2013)}]{Bache2013}
Bache K, Lichman M (2013).
\newblock \enquote{{UCI} Machine Learning Repository.}
\newblock \urlprefix\url{http://archive.ics.uci.edu/ml}.

\bibitem[{Blattenberger and Fowles(2014)}]{Blattenberger2014}
Blattenberger G, Fowles R (2014).
\newblock \enquote{Avalanche Forecasting: Using Bayesian Additive Regression
  Trees ({BART}).}
\newblock In \emph{Demand for Communications Services--Insights and
  Perspectives}, pp. 211--227. Springer-Verlag.

\bibitem[{Bleich \emph{et~al.}(2014)Bleich, Kapelner, Jensen, and
  George}]{Bleich2014}
Bleich J, Kapelner A, Jensen S, George E (2014).
\newblock \enquote{Variable Selection for BART: An Application to Gene
  Regulation.}
\newblock \emph{Annals of Applied Statistics}, \textbf{8}(3), 1750--1781.

\bibitem[{Breiman(2001)}]{Breiman2001}
Breiman L (2001).
\newblock \enquote{{Statistical Modeling: The Two Cultures}.}
\newblock \emph{Statistical Science}, \textbf{16}(3), 199--231.

\bibitem[{Chipman \emph{et~al.}(2010)Chipman, George, and
  McCulloch}]{Chipman2010}
Chipman H, George E, McCulloch R (2010).
\newblock \enquote{{BART: Bayesian Additive Regressive Trees}.}
\newblock \emph{The Annals of Applied Statistics}, \textbf{4}(1), 266--298.

\bibitem[{Chipman and McCulloch(2010)}]{BayesTree}
Chipman H, McCulloch R (2010).
\newblock \emph{BayesTree: Bayesian Methods for Tree Based Models}.
\newblock R package version 0.3-1.1,
  \urlprefix\url{http://CRAN.R-project.org/package=BayesTree}.

\bibitem[{Damien \emph{et~al.}(2013)Damien, Dellaportas, Polson, and
  Stephens}]{Damien2013}
Damien P, Dellaportas P, Polson N, Stephens D (2013).
\newblock \emph{Bayesian Theory and Applications}, pp. 455--464.
\newblock First edition. Oxford University Press.

\bibitem[{Ding and Simonoff(2010)}]{Ding2010}
Ding Y, Simonoff J (2010).
\newblock \enquote{{An Investigation of Missing Data Methods for Classification
  Trees Applied to Binary Response Data}.}
\newblock \emph{The Journal of Machine Learning Research}, \textbf{11},
  131--170.

\bibitem[{Eliashberg(2010)}]{Eliashberg2010}
Eliashberg J (2010).
\newblock \emph{Green-lighting Movie Scripts: Revenue Forecasting and Risk
  Management}.
\newblock Ph.D. thesis, University of Pennsylvania.

\bibitem[{Friedman(1991)}]{Friedman1991}
Friedman J (1991).
\newblock \enquote{{Multivariate Adaptive Regression Splines}.}
\newblock \emph{The annals of statistics}, \textbf{19}, 1--67.

\bibitem[{Friedman(2001)}]{Friedman2001}
Friedman J (2001).
\newblock \enquote{{Greedy Function Approximation: A Gradient Boosting
  Machine}.}
\newblock \emph{The Annals of Statistics}, \textbf{29}(5), 1189--1232.

\bibitem[{Friedman(2002)}]{Friedman2002}
Friedman J (2002).
\newblock \enquote{{Stochastic Gradient Boosting}.}
\newblock \emph{Computational Statistics \& Data Analysis}, \textbf{38}(4),
  367--378.

\bibitem[{Gelman \emph{et~al.}(2004)Gelman, Carlin, Stern, and
  Rubin}]{Gelman2004}
Gelman A, Carlin J, Stern H, Rubin D (2004).
\newblock \emph{Bayesian Data Analysis}.
\newblock Second edition. Chapman \& Hall / CRC.

\bibitem[{Geman and Geman(1984)}]{Geman1984}
Geman S, Geman D (1984).
\newblock \enquote{Stochastic Relaxation, {G}ibbs Distributions, and the
  {B}ayesian Restoration of Images.}
\newblock \emph{IEEE Transaction on Pattern Analysis and Machine Intelligence},
  \textbf{6}, 721--741.

\bibitem[{Hastie and Tibshirani(2000)}]{Hastie2000}
Hastie T, Tibshirani R (2000).
\newblock \enquote{{Bayesian Backfitting}.}
\newblock \emph{Statistical Science}, \textbf{15}(3), 196--213.

\bibitem[{Hastings(1970)}]{Hastings1970}
Hastings W (1970).
\newblock \enquote{{Monte Carlo Sampling Methods using Markov Chains and their
  Applications}.}
\newblock \emph{Biometrika}, \textbf{57}(1), 97--109.

\bibitem[{Kapelner and Bleich(2014)}]{Kapelner2013}
Kapelner A, Bleich J (2014).
\newblock \enquote{Prediction with Missing Data via Bayesian Additive
  Regression Trees.}
\newblock \emph{ArXiv e-prints}.

\bibitem[{Kibler \emph{et~al.}(1989)Kibler, Aha, and Albert}]{Kibler1989}
Kibler D, Aha D, Albert M (1989).
\newblock \enquote{Instance Based Prediction of Real Valued Attributes.}
\newblock \emph{Computational Intelligence}, \textbf{5}, 51.

\bibitem[{Kindo \emph{et~al.}(2013)Kindo, Wang, and Pe}]{Kindo2013}
Kindo B, Wang H, Pe E (2013).
\newblock \enquote{{MBACT - Multiclass Bayesian Additive Classification
  Trees}.}
\newblock pp. 1--29.
\newblock \eprint{arXiv:1309.7821v1}.

\bibitem[{Liaw and Wiener(2002)}]{Liaw2002}
Liaw A, Wiener M (2002).
\newblock \enquote{Classification and Regression by randomForest.}
\newblock \emph{R News}, \textbf{2}(3), 18--22.

\bibitem[{Pratola \emph{et~al.}(2013)Pratola, Chipman, Higdon, McCulloch, and
  Rust}]{Pratola2013}
Pratola M, Chipman H, Higdon D, McCulloch R, Rust W (2013).
\newblock \enquote{Parallel Bayesian Additive Regression Trees.}
\newblock \emph{Technical report}, University of Chicago.

\bibitem[{{R Core Team}(2014)}]{Rlang}
{R Core Team} (2014).
\newblock \emph{R: A Language and Environment for Statistical Computing}.
\newblock R Foundation for Statistical Computing, Vienna, Austria.
\newblock \urlprefix\url{http://www.R-project.org/}.

\bibitem[{Taddy \emph{et~al.}(2011)Taddy, Gramacy, and Polson}]{Taddy2011}
Taddy M, Gramacy R, Polson N (2011).
\newblock \enquote{{Dynamic Trees for Learning and Design}.}
\newblock \emph{Journal of the American Statistical Association},
  \textbf{106}(493), 109--123.

\bibitem[{Twala \emph{et~al.}(2008)Twala, Jones, and Hand}]{Twala2008}
Twala B, Jones M, Hand D (2008).
\newblock \enquote{{Good Methods for Coping with Missing Data in Decision
  Trees}.}
\newblock \emph{Pattern Recognition Letters}, \textbf{29}(7), 950--956.

\bibitem[{Urbanek(2013)}]{rJava}
Urbanek S (2013).
\newblock \emph{rJava: Low-level R to Java interface}.
\newblock R package version 0.9-6,
  \urlprefix\url{http://CRAN.R-project.org/package=rJava}.

\bibitem[{Zhou and Liu(2008)}]{Zhou2008}
Zhou Q, Liu JS (2008).
\newblock \enquote{Extracting Sequence Features to Predict Protein--{DNA}
  Interactions: a Comparative Study.}
\newblock \emph{Nucleic acids research}, \textbf{36}(12), 4137--4148.

\end{thebibliography}

\appendix
\section{Sampling to Modify Tree Structure}\label{app:implementation}

This section provides details on the implementation of Equation~\ref{eq:gibbs_sampler} (steps 1, 3, \ldots, $2m-1$), the Metropolis-Hastings step for sampling new trees.  Recall from Section~\ref{subsec:posterior} that trees can be altered via growing new child nodes from an existing terminal node, pruning two terminal nodes such that their parent becomes terminal, or changing the splitting rule in a node. 

Below is the Metropolis ratio \citep[p.291]{Gelman2004} where the parameter sampled is the tree and the data is the responses unexplained by other trees denoted by $\R$. We denote the new, proposal tree with an asterisk and the original tree without the asterisk.

\bneqn\label{eq:naive_metropolis}
r = \frac{\prob{\treet{*} \rightarrow \treet{}}}{\prob{\treet{} \rightarrow \treet{*}}} \frac{\cprob{\treet{*}}{\R, \sigsq}}{\cprob{\treet{}}{\R, \sigsq}}
\eneqn

We accept a draw from the posterior distribution of trees if a draw from a standard uniform distribution is less than the value of $r$. Immediately we note that it is difficult (if not impossible) to calculate the posterior probabilities for the trees themselves. Instead, we employ Bayes' Rule, 

\beqn
\cprob{\treet{}}{\R, \sigsq} = \frac{\cprob{\R}{\treet{}, \sigsq} \cprob{\treet{}}{\sigsq}}{\cprob{\R}{\sigsq}},
\eeqn

and plug the result into Equation~\ref{eq:naive_metropolis} to obtain:

\beqn
r &=& \underbrace{\frac{\prob{\treet{*} \rightarrow \treet{}}}{\prob{\treet{} \rightarrow \treet{*}}}}_{\text{transition ratio}} ~~~\times~~~ \underbrace{\frac{\cprob{\R}{\treet{*}, \sigsq}}{\cprob{\R}{\treet{}, \sigsq}}}_{\text{likelihood ratio}} ~~~\times \underbrace{\frac{\prob{\treet{*}}}{\prob{\treet{}}}}_{\text{tree structure ratio}}.
\eeqn

\noindent Note that the probability of the tree structure is independent of $\sigsq$. 

The goal of this section is to explicitly calculate $r$ for all possible tree proposals --- GROW, PRUNE and CHANGE. For each proposal, the calculations are organized into separate sections detailing each of the three ratios --- transition, likelihood and tree structure. Note that our actual implementation uses the following expressions in log form for numerical accuracy.

\subsection{Grow proposal}\label{subapp:grow_step}

\subsubsection*{Transition ratio}

Transitioning from the original tree to a new tree involves growing two child nodes from a current terminal node:

\bneqn\label{eq:grow_transition}
\prob{\treet{} \rightarrow \treet{*}} &=& \prob{\text{GROW}} \prob{\text{selecting $\eta$ to grow from}} \times \\ \nonumber
&& \prob{\text{selecting the $j$th attribute to split on}} \times \\ \nonumber
&& \prob{\text{selecting the $i$th value to split on}} \\ \nonumber
&=& \prob{\text{GROW}} \oneover{b} \oneover{\padjeta} \frac{1}{\nadjeta}.
\eneqn

We chose one of the current $b$ terminal nodes which we denote the $\eta$th node, and then we pick an attribute and split point. $\padjeta$ denotes the number of predictors left available to split on. This can be less than $p$ if certain predictors do not have two or more unique values once the data reaches the $\eta$th node. For example, this regularly occurs if a dummy variable was split on in some node higher up in the lineage. $\nadjeta$ denotes the number of \textit{unique} values left in the $p$th attribute after adjusting for parents' splits.

Transitioning from the new tree back to the original tree involves pruning that node:

\beqn
\prob{\treet{*} \rightarrow \treet{}} = \prob{\text{PRUNE}} \prob{\text{selecting $\eta$ to prune from}} =  \prob{\text{PRUNE}}\oneover{w_2^*}
\eeqn

\noindent where $w_2^*$ denotes the number of second generation internal nodes (nodes with two terminal child nodes) in the new tree. Thus, the full transition ratio is:

\beqn
\frac{\prob{\treet{*} \rightarrow \treet{}}}{\prob{\treet{} \rightarrow \treet{*}}} = \frac{\prob{\text{PRUNE}}}{\prob{\text{GROW}}} \frac{b ~ \padjeta ~ \nadjeta}{w_2^*}.
\eeqn

Note that when there are no variables with more two or more unique values, the probability of GROW is set to zero and the step will be automatically rejected.

\subsubsection*{Likelihood ratio}

To calculate the likelihood, the tree structure determines which responses fall into which of the $b$ terminal nodes. Thus,

\beqn
\cprob{\Roneton}{\treet{}, \sigsq} = \prod_{\ell=1}^{b} \cprob{\Rlonetonl}{\sigsq}
\eeqn

\noindent where each term on the right hand side is the probability of responses in one of the $b$ terminal nodes, which are independent by assumption. The $R_\ell$'s denote the data in the $\ell$th terminal node and where $n_\ell$ denotes how many observations are in each terminal node and $n = \sum_{\ell=1}^b n_\ell$. 

We now find an analytic expression for the node likelihood term. Remember, if the mean in each terminal node, which we denote $\mu_\ell$, was known, then we would have \\ $\Rlonetonl \,|\, \mu_\ell, \sigsq ~\iid~ \normnot{\mu_\ell}{\sigsq}$. BART requires $\mu_\ell$ to be integrated out, allowing the Gibbs sampler in Equation~\ref{eq:gibbs_sampler} to avoid dealing with jumping between continuous spaces of varying dimensions \citep[page 275]{Chipman2010}. Recall that one of the BART model assumptions is a prior on the average value of $\mu \sim \normnot{0}{\sigsq_\mu}$ and thus,

\beqn
\cprob{\Rlonetonl}{\sigsq} = \int_\reals \cprob{\Rlonetonl}{\mu_\ell, \sigsq} \prob{\mu_\ell; \sigsq_\mu} d\mu_\ell
\eeqn

\noindent which can be shown via completion of the square or convolution to be

\bneqn\label{eq:likelihood_margined}
\cprob{\Rlonetonl}{\sigsq} &=& \oneover{\tothepow{2\pi\sigsq}{n_\ell / 2}} \sqrt{ \frac{\sigsq}{\sigsq + n_\ell \sigsqmu}}  \times \\ 
&& \exp{-\oneover{2\sigsq} \parens{\sum_{i=1}^{n_\ell} \squared{R_{\ell_i} - \Rbar_\ell} - \frac{\Rbar_\ell^2 n_\ell^2}{n_\ell + \frac{\sigsq}{\sigsqmu}}+ n_\ell \Rbar_\ell^2}} \nonumber
\eneqn

\noindent where $\Rbar_\ell$ denotes the mean response in the node and $R_{\ell_i}$ denotes the observations $i=1\ldots n_\ell$ in the node.

Since the likelihoods are solely determined by the terminal nodes, the proposal tree differs from the original tree by only the selected node to be grown, denoted by $\ell$, which becomes two children after the GROW step denoted by $\ell_L$ and $\ell_R$. Hence, the likelihood ratio becomes:

\bneqn\label{eq:grow_likelihood_ratio}
&& \frac{\cprob{\R}{\treet{*}, \sigsq}}{\cprob{\R}{\treet{}, \sigsq}} = \frac{\cprob{\RLlonetonlL}{\sigsq} \cprob{\RRlonetonlR}{\sigsq}}{\cprob{\Rlonetonl}{\sigsq}} 
\eneqn

Plugging Equation~\ref{eq:likelihood_margined} into Equation~\ref{eq:grow_likelihood_ratio} three times yields the ratio for the GROW step:

\small
\beqn
 \sqrt{\frac{\sigsq \parens{\sigsq + n_\ell \sigsqmu}}{\parens{\sigsq + n_{\ell_L} \sigsqmu}\parens{\sigsq + n_{\ell_R} \sigsqmu}}}~~\exp{\frac{\sigsq_\mu}{2\sigsq} \parens{\frac{\squared{\sum_{i=1}^{n_{\ell_L}} R_{\ell_L, i}}}{\sigsq + n_{\ell_L}\sigsq_\mu} + \frac{\squared{\sum_{i=1}^{n_{\ell_R}} R_{\ell_R, i}}}{\sigsq + n_{\ell_R}\sigsq_\mu} - \frac{\squared{\sum_{i=1}^{n_{\ell}} R_{\ell, i}}}{\sigsq + n_\ell \sigsq_\mu}}} \\
\eeqn
\normalsize

\noindent where $n_{\ell_L}$ and $n_{\ell_R}$ denote the number of data points in the newly grown left and right child nodes.

\subsubsection*{Tree structure ratio}

In Section~\ref{subsec:prior_likelihood} we discussed the prior on the tree structure (where the splits occur) as well as the tree rules. For the entire tree,

\beqn
\prob{\treet{}} &=& \prod_{\eta \in \Hterminals} \parens{1 - \probsplit{\eta}}  \prod_{\eta \in \Hint} \probsplit{\eta} \prod_{\eta \in \Hint} \probrule{\eta} \\
\eeqn

\noindent where $\Hterminals$ denotes the set of terminal nodes and $\Hint$ denotes the internal nodes.

Recall that the probability of splitting on a given node $\eta$ is $\probsplit{\eta} = \alpha / \tothepow{1 + d_\eta}{\beta}$. The probability is controlled by two hyperparameters, $\alpha$ and $\beta$, and $d_\eta$ is the depth (number of parent generations) of node $\eta$. When assigning a rule, recall that BART picks from all available attributes and then from all available unique split points. Using the notation from the transition ratio section, $\probrule{\eta} = 1 / \padjeta \times 1 / \nadjeta$.

Once again, the original tree features a node $\eta$ that was selected to be grown. The proposal tree differs with two child nodes denoted $\eta_L$ and $\eta_R$. We can now form the ratio:

\beqn
\frac{\prob{\treet{*}}}{\prob{\treet{}}} &=& \frac{\parens{1 - \probsplit{\eta_L}} \parens{1 - \probsplit{\eta_R}} \probsplit{\eta} \probrule{\eta}}{\parens{1 - \probsplit{\eta}}}\\
&=& \frac{\parens{1 - \dfrac{\alpha}{\tothepow{1 + d_{\eta_L}}{\beta}}}\parens{1 - \dfrac{\alpha}{\tothepow{1 + d_{\eta_R}}{\beta}}} \dfrac{\alpha}{\tothepow{1 + d_{\eta}}{\beta}} \doneover{\padjeta}\dfrac{1}{\nadjeta}}{1 -\frac{\alpha}{\tothepow{1 + d_{\eta}}{\beta}}} \\
&=& \alpha \frac{\squared{1 - \frac{\alpha}{\tothepow{2 + d_\eta}{\beta}}}}{\parens{\tothepow{1+d_\eta}{\beta} - \alpha} \padjeta \nadjeta}
\eeqn

The last line follows from algebra and using the fact that the depth of the grown nodes is the depth of the parent node incremented by one ($d_{\eta_L} = d_{\eta_R} = d_{\eta} + 1$).

\subsection{Prune proposal}\label{subapp:prune_step}

A prune proposal is the \qu{opposite} of a grow proposal. Prune selects a singly internal node (a node whose children are both terminal) and removes both of its children. Thus, each ratio will be approximately the inverse of the ratios found in the previous section concerning the grow proposal. Note also that prune steps are not considered in trees that consist of a single root node.

\subsubsection*{Transition ratio}

We begin with transitioning from the original tree to the proposal tree:

\beqn
\prob{\treet{} \rightarrow \treet{*}} = \prob{\text{PRUNE}} \prob{\text{selecting $\eta$ to prune from}} =  \prob{\text{PRUNE}}\oneover{w_2}
\eeqn

\noindent where $w_2$ denotes the number of singly internal parent nodes which have two terminal children (thus no grandchildren). To transition in the opposite direction, we are obligated to grow from node $\eta$. This is similar to Equation~\ref{eq:grow_transition} except the proposed tree has one less terminal node due to the pruning of the original tree, resulting in a  $1 / (b-1)$ term:

\beqn
\prob{\treet{*} \rightarrow \treet{}} &=&  \prob{\text{GROW}} \oneover{b-1} \oneover{\padjetastar} \frac{1}{\nadjetastar}.
\eeqn

Thus, the transition ratio is:

\beqn
\frac{\prob{\treet{*} \rightarrow \treet{}}}{\prob{\treet{} \rightarrow \treet{*}}} = \frac{\prob{\text{GROW}}}{\prob{\text{PRUNE}}} \frac{w_2 }{(b-1) \padjetastar \nadjetastar}.
\eeqn

\subsubsection*{Likelihood ratio}

This is simply the inverse of the likelihood ratio for the grow proposal:

\beqn
\frac{\cprob{\R}{\treet{*}, \sigsq}}{\cprob{\R}{\treet{}, \sigsq}} &=&  \sqrt{\frac{\parens{\sigsq + n_{\ell_L} \sigsqmu}\parens{\sigsq + n_{\ell_R} \sigsqmu}}{\sigsq \parens{\sigsq + n_\ell \sigsqmu}}} \times \\
&& \exp{\frac{\sigsq_\mu}{2\sigsq} \parens{\frac{\squared{\sum_{i=1}^{n_{\ell_{~}}} R_{\ell, i}}}{\sigsq + n_\ell \sigsq_\mu} - \frac{\squared{\sum_{i=1}^{n_{\ell_L}} R_{\ell_L, i}}}{\sigsq + n_{\ell_L}\sigsq_\mu} - \frac{\squared{\sum_{i=1}^{n_{\ell_R}} R_{\ell_R, i}}}{\sigsq + n_{\ell_R}\sigsq_\mu}}}. \\
\eeqn

\subsubsection*{Tree structure ratio}

This is also simply the inverse of the tree structure ratio for the grow proposal:

\beqn
\frac{\prob{\treet{*}}}{\prob{\treet{}}}  &=& \frac{\parens{\tothepow{1+d_\eta}{\beta} - \alpha} \padjetastar \nadjetastar}{\alpha\squared{1 - \frac{\alpha}{\tothepow{2 + d_\eta}{\beta}}}}.
\eeqn

\subsection{Change proposal}\label{subapp:change_step}

A change proposal involves picking an internal node and changing its rule by picking both a new available predictor to split on and a new valid split value among values of the selected predictor.  Although this could be implemented for use in any internal node in the tree, for simplicity we limit our implementation to singly internal nodes: those that have two terminal child nodes.

\subsubsection*{Transition ratio}

The transition to a proposal tree is below:

\beqn
\prob{\treet{} \rightarrow \treet{*}} &=& \prob{\text{CHANGE}} \prob{\text{selecting node $\eta$ to change}} \times \\
&& \prob{\text{selecting the new attribute to split on}} \times \\
&& \prob{\text{selecting the new value to split on}}
\eeqn

When calculating the ratio, the first three terms are shared in both numerator and denominator. The probability of selecting the new value to split on will differ as different split features have different numbers of unique values available. Thus we are left with

\beqn
\frac{\prob{\treet{*} \rightarrow \treet{}}}{\prob{\treet{} \rightarrow \treet{*}}} = \frac{\nadjetastar}{\nadjeta}
\eeqn

\noindent where $\nadjetastar$ is the number of split values available under the proposal tree's splitting rule and $\nadjeta$ is the number of split values available under the original tree's splitting rule.

\subsubsection*{Likelihood ratio}

The proposal tree differs from the original tree only in the two child nodes of the selected change node. These two terminal nodes have the unexplained responses apportioned differently. Denote $R_{1\cdot}$ as the residuals of the first child node and $R_{2\cdot}$ as the unexplained responses in the second child node. Thus we begin with:

\beqn
\frac{\cprob{\R}{\treet{*}, \sigsq}}{\cprob{\R}{\treet{}, \sigsq}} &=& \frac{\cprob{\Ronestars}{\sigsq}\cprob{\Rtwostars}{\sigsq}}{\cprob{\Rones}{\sigsq}\cprob{\Rtwos}{\sigsq}}
\eeqn

\noindent where the responses denoted with an asterisk are the responses in the proposal tree's child nodes.

Substituting Equation~\ref{eq:likelihood_margined} four times and using algebra, the following expression is obtained for the ratio:

\beqn
&& \sqrt{\frac{\parens{\frac{\sigsq}{\sigsqmu} + n_1} \parens{\frac{\sigsq}{\sigsqmu} + n_2}}{\parens{\frac{\sigsq}{\sigsqmu} + n_1^*} \parens{\frac{\sigsq}{\sigsqmu} + n_2^*}}} \times \\
&&  \exp{\oneover{2\sigsq} \parens{\frac{\squared{\sum_{i=1}^{\nonestar} R_{1^*, i}}}{\nonestar + \frac{\sigsq}{\sigsqmu}} + \frac{\squared{\sum_{i=1}^{\ntwostar} R_{2^*, i}}}{\ntwostar + \frac{\sigsq}{\sigsqmu}} - \frac{\squared{\sum_{i=1}^{\none} R_{1, i}}}{\none + \frac{\sigsq}{\sigsqmu}} - \frac{\squared{\sum_{i=1}^{\ntwo} R_{2, i}}}{\ntwo + \frac{\sigsq}{\sigsqmu}}}}
\eeqn

\noindent which simplifies to

\beqn
\exp{\oneover{2\sigsq} \parens{\frac{\squared{\sum_{i=1}^{\nonestar} R_{1^*, i}} - \squared{\sum_{i=1}^{\nonestar} R_{1, i}}}{\none + \frac{\sigsq}{\sigsqmu}} + \frac{\squared{\sum_{i=1}^{\nonestar} R_{2^*, i}} - \squared{\sum_{i=1}^{\nonestar} R_{2, i}}}{\ntwo + \frac{\sigsq}{\sigsqmu}}}}
\eeqn

\noindent if the number of responses in the children do not change in the proposal ($n_1 = n_1^*$ and $n_2 = n_2^*$).

\subsubsection*{Tree structure ratio}

The proposal tree has the same structure as the original tree. Thus we only need to take into account the changed node's children:

\beqn
\frac{\prob{\treet{*}}}{\prob{\treet{}}} = \frac{\parens{1 - \probsplit{\etaonestar}} \parens{1 - \probsplit{\etatwostar}} \probsplit{\etastar} \probrule{\etastar}}{\parens{1 - \probsplit{\etaone} \parens{1 - \probsplit{\etatwo}}} \probsplit{\eta} \probrule{\eta}}.
\eeqn

The probability of splits remain the same because the child nodes are at the same depths. Thus we only need to consider the ratio of the probability of the rules. Once again, the probability of selecting the new value to split on will differ as different split features have different numbers of unique values available. We are left with $\mathbb{P}(\treet{*}) / \mathbb{P}(\treet{}) = \nadjeta / \nadjetastar$.

Note that this is the inverse of the transition ratio. Hence, for the change step, only the likelihood ratio needs to be computed to determine the Metropolis-Hastings ratio $r$.

\section{Bakeoff}\label{app:bakeoff}

We baked off nine regression data sets and assessed out-of-fold RMSE using 10-fold cross-validation. We average the results across 20 replications of cross-validation. The results are displayed in Table~\ref{tab:bakeoff}.

\begin{table}[htp]
\centering
\begin{tabular}{rrrr}
  \hline
 & bartMachine & BayesTree & RF \\ 
  \hline
boston & 4.451~\, & 4.503~\, & 4.582 \\ 
  triazine & 0.128* & 0.130~\, & 0.119 \\ 
  ozone & 4.147~\, & 4.144~\, & 4.064 \\ 
  baseball & 709.197~\, & 709.437~\, & 729.188 \\ 
  wine.red & 0.656~\, & 0.651* & 0.642 \\ 
  ankara & 1.348* & 1.461~\, & 1.574 \\ 
  wine.white & 0.759* & 0.766~\, & 0.746 \\ 
  pole & 11.713* & 12.755~\, & 10.691 \\ 
  compactiv & 3.262~\, & 2.795* & 2.957 \\ 
   \hline
\end{tabular}
\caption{RMSE values for three machine learning algorithms averaged across 20 replicates. Asterisks indicate a significant difference between \pkg{bartMachine} and \pkg{BayesTree} at a significance level of 5\% with a Bonferroni correction. Comparisons with \pkg{randomForest}'s performance were not conducted.}
\label{tab:bakeoff}
\end{table}

We conclude that the implementation outlined in this paper performs approximately the same as the previous implementation with regards to predictive accuracy.

Table~\ref{tab:times} shows the average run-time for each algorithm. Note that \proglang{bartMachine} is run using 4 cores. 

\begin{table}[htp]
\centering
\begin{tabular}{rrrr}
  \hline
 & bartMachine & BayesTree & RF \\ 
  \hline
boston & 7.8 & 28.5 & 5.1 \\ 
  triazine & 5.7 & 10.7 & 2.6 \\ 
  ozone & 4.7 & 17.6 & 2.1 \\ 
  baseball & 5.6 & 18.6 & 3.3 \\ 
  wine.red & 13.5 & 51.1 & 10.6 \\ 
  ankara & 12.8 & 27.0 & 10.9 \\ 
  wine.white & 13.5 & 56.0 & 11.0 \\ 
  pole & 18.2 & 7.0 & 12.0 \\ 
  compactiv & 16.3 & 18.4 & 19.2 \\ 
   \hline
\end{tabular}
\caption{Average run-times (in seconds) for each complete k-fold estimation for three machine learning algorithms.}
\label{tab:times}
\end{table}

\end{document}